\documentclass[journal,twoside,web]{ieeecolor}

\usepackage{tmi}
\usepackage{cite}
\usepackage{amsmath,amssymb,amsfonts}
\usepackage{algorithmic}
\usepackage{graphicx}
\usepackage{textcomp}
\usepackage{xcolor}
\usepackage{multirow}
\usepackage{tabu}
\usepackage{svg}
\usepackage{bbm}
\usepackage{comment}
\usepackage{array}
\usepackage{amstext}
\usepackage{makecell}
\usepackage{caption}
\usepackage{tablefootnote}
\usepackage{booktabs} 
\usepackage{float}
\usepackage{multicol}
\usepackage{lipsum}
\usepackage{cuted}

\usepackage{colortbl}

\usepackage{placeins}
\usepackage{nicefrac} 

\newcommand{\ie}{\textit{i.e.}}
\newcommand{\eg}{\textit{e.g.}}

\usepackage[pagebackref=false,breaklinks=true,colorlinks,bookmarks=false]{hyperref}

\def\BibTeX{{\rm B\kern-.05em{\sc i\kern-.025em b}\kern-.08em
T\kern-.1667em\lower.7ex\hbox{E}\kern-.125emX}}
\newcommand{\bfsection}[1]{\vspace*{0.1cm}\noindent\textbf{#1.}}
\newcommand{\itsection}[1]{\noindent\textit{#1.}}

\newcommand{\revised}[1]{{#1}}
\newcommand{\revisedsecond}[1]{{#1}}

\begin{document}
\title{Frenet-Serret Frame-based Decomposition for Part Segmentation of 3D Curvilinear Structures}

\author{Leslie Gu, Jason Ken Adhinarta, Mikhail Bessmeltsev, Jiancheng Yang, Yongjie Jessica Zhang, \\Wenjie Yin, Daniel Berger, Jeff Lichtman, Hanspeter Pfister, Donglai Wei
\thanks{This work was partially supported by NSF grant NCS-FO-2124179, and NIH grant 1U01NS132158. Leslie Gu is also supported by SEAS Graduate Fellowship. Donglai Wei is supported by NSF-IIS 2239688.}
\thanks{Shixuan Leslie Gu and Hanspeter Pfister are with the John A. Paulson School of Engineering and Applied Sciences, Harvard University, MA, USA (\{shixuangu@g,pfister@seas\}.harvard.edu).} 
\thanks{Jason Ken Adhinarta and Donglai Wei are with Boston College, MA, USA (\{jason.adhinarta,donglai.wei\}@bc.edu).}
\thanks{Mikhail Bessmeltsev is with Université de Montréal, QC, Canada (bmpix@iro.umontreal.ca).}
\thanks{Jiancheng Yang is with 
Swiss Federal Institute of Technology Lausanne, Lausanne, Switzerland (jiancheng.yang@epfl.ch).}
\thanks{Yongjie Jessica Zhang is with Carnegie Mellon University, PA, USA (jessicaz@andrew.cmu.edu).}
\thanks{Wenjie Yin, Daniel Berger, and Jeff Lichtman are with the Department of Molecular and Cellular Biology, Harvard University, MA, USA (kelly.wjyin@gmail.com, \{danielberger@fas,jlichtman@mcb\}.harvard.edu).} 
}

\maketitle

\begin{abstract}
\revisedsecond{Accurate segmentation of anatomical substructures within 3D curvilinear structures} in medical imaging remains challenging due to their complex geometry and the scarcity of diverse, large-scale datasets for algorithm development and evaluation. In this paper, we use dendritic spine segmentation as a case study and address these challenges by introducing a novel Frenet–Serret Frame-based Decomposition, which decomposes 3D curvilinear structures into a globally smooth continuous curve that captures the overall shape, and a cylindrical primitive that encodes local geometric properties. This approach leverages Frenet–Serret Frames and arc length parameterization to preserve essential geometric features while reducing representational complexity, facilitating data-efficient learning, improved segmentation accuracy, and generalization on 3D curvilinear structures. To rigorously evaluate our method, we introduce two datasets: \textit{CurviSeg}, a synthetic dataset for 3D curvilinear structure segmentation that validates our method's key properties, and \textit{DenSpineEM}, a benchmark for dendritic spine segmentation, which comprises 4,476 manually annotated spines from 70 dendrites across three public electron microscopy datasets, covering multiple brain regions and species. Our experiments on \textit{DenSpineEM} demonstrate exceptional cross-region and cross-species generalization:
models trained on the mouse somatosensory cortex subset achieve \revised{94.43\%} Dice, maintaining strong performance in zero-shot segmentation on both mouse visual cortex (\revised{95.61\%} Dice) and human frontal lobe (\revised{86.63\%} Dice) subsets. Moreover, we test the generalizability of our method on the \textit{IntrA} dataset, where it achieves 77.08\% Dice (5.29\% higher than prior arts) on intracranial aneurysm segmentation \revised{from entire artery models}. These findings demonstrate the potential of our approach for accurately analyzing complex curvilinear structures across diverse medical imaging fields. Our dataset, code, and models are available at \url{https://github.com/VCG/FFD4DenSpineEM} to support future research.
\end{abstract}



\begin{figure}[!t]
\centering
\includegraphics[width=\linewidth]{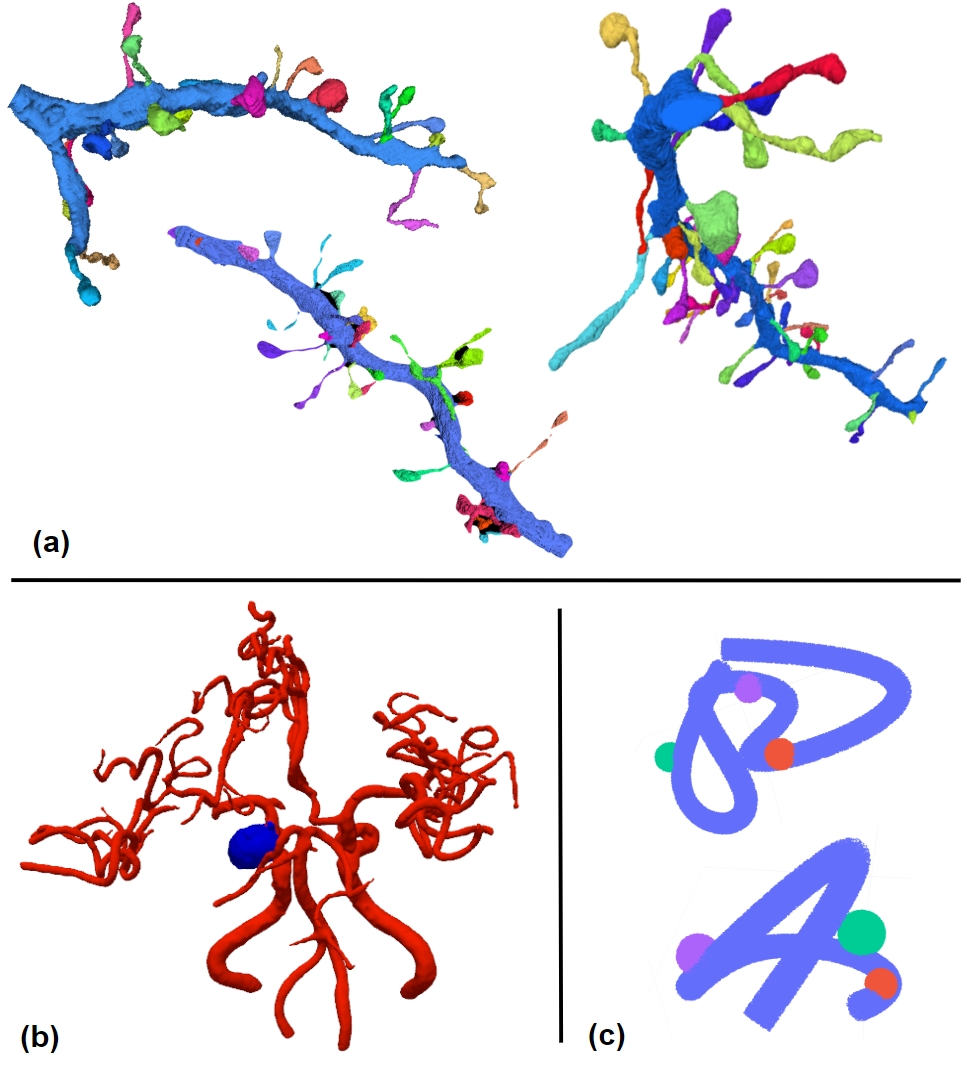}
\caption{\textbf{Part Segmentation for 3D Curvilinear Structures.} Curvilinear structures from (a) DenSpineEM: Main experimental dataset on dendritic spine segmentation. (b) IntrA: Intracranial aneurysm segmentation dataset for testing cross-domain generalizability. (c) CurviSeg: Synthetic dataset for theoretical validation. Colors indicate segmentation labels.} \label{fig:dataset_overview}
\vspace{-1.5em}
\end{figure}



\section{Introduction}
\label{sec:introduction}
\IEEEPARstart{D}{eep} learning-enabled 3D biomedical imaging has driven advancements in both scientific research (\eg, connectomics~\cite{Fornito2015TheCO,Wu2023MappingON}, protein structure prediction~\cite{Jumper2021HighlyAP,Abramson2024AccurateSP}) and as a crucial tool in medical care (\eg, bone lesion analysis~\cite{yang2021ribseg,Jin2022RibSeg,Yang2024DeepRF}, aneurysm detection~\cite{Yang2020IntrA3I}). While semantic segmentation algorithms, such as nn-UNet~\cite{Isensee2020nnUNetAS}, have achieved strong results in various tasks, the segmentation of 3D curvilinear structures remains challenging due to their intricate geometry, varying thickness, and complex branching patterns~\cite{Kashiwagi2019ComputationalGA}. These structures, characterized by their elongated, often branching nature following curved paths in three-dimensional space, are ubiquitous in biological and medical imaging, playing crucial roles in various systems from neuronal networks to vascular systems~\cite{Boros2017DendriticSP,Qian2023BiomimeticIN}.



In this paper, we focus on dendritic spine segmentation as a representative task for 3D curvilinear structure analysis. Dendritic spines, small protrusions on neuronal dendrites, are crucial for synaptic transmission, and their morphology and density provide vital information about neuronal connectivity, making accurate segmentation essential for neuroscience research~\cite{Boros2017DendriticSP}. However, segmentation is challenging due to spines' high density along dendrites, complex geometry, variable sizes and shapes, and intricate branching patterns~\cite{Kashiwagi2019ComputationalGA}. The lack of benchmark datasets has led to reliance on simple heuristics without human-annotated comparisons, limiting the reliability of current methods.

Recent advances, such as deep learning-based workflows~\cite{VidaurreGallart2022ADL}, joint classification and segmentation methods for 2-photon microscopy images~\cite{Erdil2015AJC}, and interactive tools like 3dSpAn~\cite{Basu20193dSpAnAI}, have improved performance. However, these approaches often require large training datasets or manual refinement and struggle to generalize across different imaging conditions and spine morphologies. This underscores the need for more data-efficient methods capable of handling the complexity of 3D curvilinear structures.


To address these challenges, we propose the \textit{Frenet–Serret Frame-based Decomposition (FFD)}, which decomposes 3D curvilinear geometries into two components: a globally smooth $C^2$ continuous curve that captures the overall shape, and a cylindrical primitive that encodes local geometric properties. This approach leverages Frenet–Serret Frames and arc length parameterization to preserve essential geometric features while reducing representational complexity. The resultant cylindrical representation facilitates data-efficient learning, improved segmentation accuracy, and generalization on 3D curvilinear structures.

To validate the effectiveness of our approach, we introduce \textit{CurviSeg}, a synthetic dataset for segmentation tasks of 3D curvilinear structures, which serves as a theoretical validation to verify the key properties of our method. Additionally, we present \textit{DenSpineEM}, a benchmark dataset for dendritic spine segmentation, consisting of 4,476 manually annotated dendritic spines from 70 dendrites across three 3D electron microscopy (EM) image stacks (mouse somatosensory cortex, mouse visual cortex, and human frontal lobe). Using our decomposition, models trained on the large subset from the mouse somatosensory cortex achieve high segmentation performance (\revised{94.43\%} Dice) and demonstrate strong zero-shot generalization on both the mouse visual cortex (\revised{95.61\%} Dice) and human frontal lobe (\revised{86.63\%} Dice) subsets. Moreover, we demonstrate the generalizability of our method on the \textit{IntrA} dataset for intracranial aneurysm segmentation \revised{from entire artery}, where it achieves 77.08\% DSC, outperforming the state-of-the-art by 5.29\%, highlighting its effectiveness beyond dendritic spine segmentation to other medical imaging tasks.

Our contributions include:
\begin{itemize}
    \item We propose the Frenet–Serret Frame-based Decomposition, decomposing 3D curvilinear geometries into a smooth $C^2$ curve\footnote{\revised{$C^2$ continuity refers to a curve that is twice continuously differentiable, meaning the curve has continuous first and second derivatives. This property ensures smoothness in both the curve and its rate of change.}} and cylindrical primitive for efficient learning and robust segmentation.
    \item We develop \textit{DenSpineEM}, a comprehensive benchmark for 3D dendritic spine segmentation, containing 4,476 manually annotated spines from 70 dendrites across three EM datasets, covering various brain regions and species.
    \item We introduce \textit{CurviSeg}, a synthetic dataset for 3D curvilinear structure segmentation, used to validate our method and as a resource for other analyses.
    \item Our method achieves high segmentation accuracy with cross-species and cross-region generalization on dendritic spine segmentation, and surpasses state-of-the-art methods on intracranial aneurysm segmentation.
\end{itemize}


\section{Related Works}

\subsection{\revised{3D Curvilinear Structure Analysis in Biomedical Imaging}}

\bfsection{\revised{Traditional Methods}}
In the medical domain, curvilinear structures are prevalent and critical, with applications spanning blood vessel segmentation~\cite{Lesage2009ARO}, neuronal tracing~\cite{Acciai2016AutomatedNT}, and airway tree extraction~\cite{Lo2012ExtractionOA}. These structures, characterized by their tubular or filament-like shape, present unique challenges due to their complex geometry and intricate branching patterns. Traditional methods rely on hand-crafted features, such as the Hessian-based Frangi vesselness filter~\cite{Frangi1998MuliscaleVE} and multi-scale line filter~\cite{Sato1998ThreedimensionalML}, which enhance tubular structures but often struggle with complex geometries and varying scales.

\bfsection{\revised{Learning-Based Approaches}}
Recent advancements leverage machine learning techniques to improve robustness and accuracy. Sironi et al. \cite{Sironi2016MultiscaleCD} introduced a multi-scale regression approach for centerline detection, while deep learning methods (e.g., nnU-Net \cite{Isensee2020nnUNetAS} and DeepVesselNet \cite{Tetteh2018DeepVesselNetVS}) have shown superior performance in vessel segmentation tasks. Despite these advances, challenges persist in the medical domain, including high variability in structure appearance, resolution limitations, and the scarcity of large-scale annotated datasets \cite{Litjens2017ASO}. Our work builds upon these foundations, using dendritic spine segmentation as a compelling example to address these challenges through our novel Frenet frame-based transformation.

\bfsection{\revised{Geometric and Topological Regularization}}
\revised{Recent research has emphasized the importance of integrating geometric and topological constraints into segmentation models for curvilinear structures. Topology-preserving approaches such as the clDice loss~\cite{Shit2020clDiceA} focus on maintaining connectivity in tubular structures by measuring the overlap between centerlines. \revisedsecond{Other methods incorporate global shape descriptors~\cite{Kervadec2021BeyondPS} to constrain segmentation beyond pixel-wise supervision or utilize centerline-based topological features to facilitate localization and segmentation tasks in vascular structures~\cite{Yao2024AASegAG}.} Geometric guidance has proven particularly effective for complex anatomical structures, as demonstrated by BowelNet~\cite{Wang2022BowelNetJS}, which integrates geometric priors with semantic segmentation for improved bowel delineation. Furthermore, discrete topology-based methods, such as topology-aware segmentation using discrete Morse theory~\cite{Hu2021TopologyAwareSU}, further emphasize the importance of preserving structural correctness during segmentation. Similarly, deep models that explicitly incorporate topological priors~\cite{Hu2019TopologyPreservingDI} have shown to effectively reduce errors like spurious holes or disconnections. These approaches highlight the benefits of combining low-level image information with high-level geometric understanding, a principle that inspires our Frenet frame-based transformation.
}

\subsection{\revised{3D Representations for Medical Imaging}}
\bfsection{\revised{Voxel-Based Representations}}
3D shapes in biomedical imaging, typically derived from CT (Computational Tomography) and EM (Electron Microscopy) scans, are often represented as voxels on discrete grids. Prior works \cite{ravishankar2017learning,wang2020deep} predominantly use voxel representations, extending 2D approaches to 3D (\eg, 3D UNet \cite{cciccek20163d}) or employing sophisticated 3D operators \cite{yang2021reinventing}. However, voxel-based methods face challenges with high memory requirements and limited spatial resolution. 

\bfsection{\revised{Point Cloud-Based Methods}}
Alternatively, point cloud representations offer a lightweight and flexible approach for 3D shape analysis \cite{qi2017pointnet}. They excel in extracting semantic information \cite{ho2021point} and provide higher computational efficiency for large-scale objects. \revised{Building on these advantages, recent works have adapted point-based deep learning frameworks for curvilinear structures, leveraging local geometric features and long-range context to capture fine anatomical detail. For instance, Liu et al.~\cite{Liu2022EdgeOrientedPT} proposed an edge-oriented point transformer for intracranial aneurysm segmentation, introducing edge-aware supervision and contrastive learning to enhance boundary delineation. Similarly, Xie et al.~\cite{Xie2023EfficientAL} presented a deep point-graph implicit field method for efficient anatomical labeling of the pulmonary airway tree, capturing topological structure while preserving spatial detail. These methods demonstrate the potential of point cloud models for fine-grained part segmentation in complex anatomical geometries. Benchmarks such as IntrA~\cite{Yang2020IntrA3I} and RibSeg~\cite{Jin2022RibSeg} further highlight the utility of point-based approaches in curvilinear structure analysis, including intracranial aneurysm and rib segmentation. However, despite their success, most existing methods focus on relatively rigid or tubular forms. In contrast, dendritic spines exhibit more diverse morphologies and branching patterns, requiring enhanced geometric awareness. Our method builds on these insights and incorporates curve-guided priors to improve segmentation of highly variable, curvilinear biological structures.}

\subsection{\revised{Dendritic Spine Segmentation}}
Dendrites, with their curvy and elongated structure, serve as an excellent example for curvilinear structure analysis. Their protrusions, known as dendritic spines, play a crucial role in neuronal connectivity and plasticity \cite{Nimchinsky2002StructureAF}. The segmentation of these spines presents unique challenges across different imaging modalities. In light microscopy, where spines appear as tiny blobs due to limited resolution, research has focused on spine location detection \cite{xiao2018automated}, semi-automatic segmentation \cite{Basu20193dSpAnAI}, and morphological analysis \cite{choi2021dxplorer}. High-resolution electron microscopy (EM) has enabled more precise spine analysis, leading to two main approaches: morphological operations with watershed propagation \cite{wildenberg2021large}, and skeletonization with radius-based classification \cite{dorkenwald2019binary}. However, these methods often rely on hand-tuned hyperparameters and require all voxels as input, limiting their effectiveness for large-scale data analysis.
The field of dendritic spine segmentation faces two significant challenges: the lack of comprehensive benchmark datasets for rigorous evaluation, and the need for effective methods that can handle complex spine geometry in large-scale datasets. To address these challenges, we introduce both a large-scale 3D dendritic spine segmentation benchmark and a novel Frenet frame-based transformation method, potentially advancing curvilinear structure analysis in neuroscience and beyond.

\subsection{\revised{Preliminaries on Frenet-Serret Frame}}
To understand the geometric properties of curvilinear structures, we turn to the fundamental concept of the Frenet-Serret frame in differential geometry. In three-dimensional Euclidean space $\mathbb{R}^{3}$, the Frenet-Serret frame (TNB frame) of a differentiable curve at a point is a triplet of three mutually orthogonal unit vectors \revised{(i.e., tangent, normal, and binormal, denoted as $\mathbf{T}$, $\mathbf{N}$, and $\mathbf{B}$, respectively.)}~\cite{Piaggio1952DifferentialGO}. Specifically, let $\mathbf{r}(s)$ be a curve in Euclidean space parameterized by arc length $\mathbf{s}$, then the Frenet-Serret frame can be defined by:
\begin{equation}
\mathbf{T}:=\frac{\mathrm{d} \mathbf{r}}{\mathrm{d} s}, ~\mathbf{N}:=\frac{\mathrm{d} \mathbf{T}}{\mathrm{d} s}/\left\|\frac{\mathrm{d} \mathbf{T}}{\mathrm{d} s}\right\|, ~\mathbf{B}:=\mathbf{T} \times \mathbf{N},
\end{equation}\label{eq:tnb}
which satisfies the Frenet-Serret formulas:
\begin{equation}
\frac{\mathrm{d}\mathbf{T}}{\mathrm{d}s} = \kappa\mathbf{N},~\frac{\mathrm{d}\mathbf{N}}{\mathrm{d}s} = -\kappa\mathbf{T} + \tau\mathbf{B},~\frac{\mathrm{d}\mathbf{B}}{\mathrm{d}s} = -\tau\mathbf{N},
\end{equation}\label{eq:frenet}
where $\kappa(s)$ is curvature and $\tau(s)$ is torsion, measuring how sharply the curve bends and how much the curve twists out of a plane. 

Originally formulated for physics applications~\cite{Bishop1975ThereIM}, Frenet-Serret Frame has subsequently been adopted across diverse domains. 
In robotics and autonomous driving, it facilitates the optimization of trajectory planning~\cite{ Huang2023TrajectoryPI}.
The computer graphics community utilizes it for generating swept surface models~\cite{Kseolu2023InvolutiveSS}, rendering streamline visualizations~\cite{Hanson1995QuaternionFA}, and computing tool paths in CAD/CAM systems~\cite{Pottmann1998ContributionsTM}. 
More recently, Frenet frame has been instrumental in characterizing protein structures in bioinformatics~\cite{Hu2011DiscreteFF}, underscoring their adaptability across varying scales and scientific disciplines. Our work extends this concept to the (bio)medical domain, specifically for the analysis and segmentation of dendritic spines, where we employ it to map these 3D curvilinear structures onto a standardized cylindrical coordinate system while preserving crucial geometric properties.

\begin{figure*}[t]
     \centering
     \includegraphics[width=\textwidth]{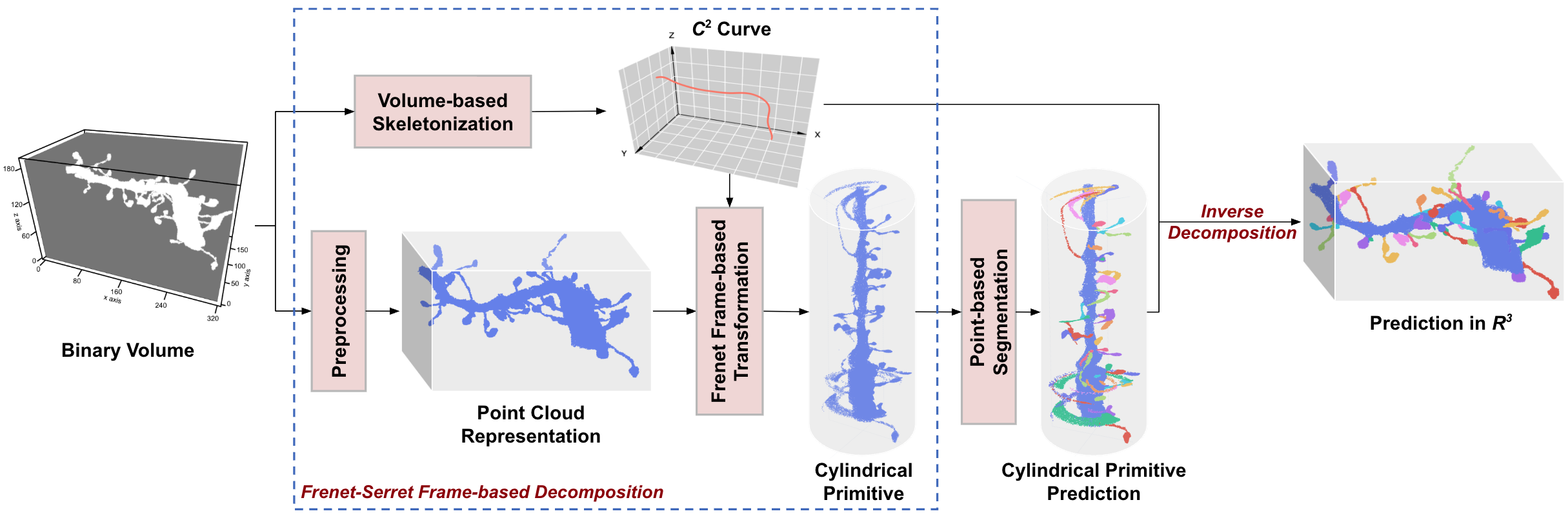}
    \caption{\textbf{Exemplary Pipeline of Dendritic Spine Segmentation using Frenet-Serret Frame-based Decomposition.} The pipeline consists of three main steps: \textbf{Decomposition:} Decomposing the input binary volume to a $C^2$ curve and a cylindrical primitive $(\mathbb{R}^+ \times S^1 \times \mathbb{R})$.    
    \revisedsecond{Specifically, 1) we first skeletonize the volume to obtain the $C^2$ curve, 2) then convert the volume to point clouds in $\mathbb{R}^3$, and 3) transform them to a cylindrical primitive in $(\mathbb{R}^+ \times S^1 \times \mathbb{R})$ using the Frenet-Serret frame of the $C^2$ curve.} \textbf{Segmentation:} Performing point-based segmentation on the cylindrical primitive, leveraging the simplified geometry for improved accuracy and efficiency. \revisedsecond{Note that the cylindrical primitive is converted to Cartesian coordinates for compatibility with point-based networks.} \textbf{Inverse Decomposition:} Reconstructing the segmented structure back to \revised{the original shape} by combining the cylindrical primitive with the $C^2$ curve.}
    
    \label{fig:teaser}
    \vspace{-1em}
\end{figure*}

\section{Frenet–Serret Frame-based Decomposition}
\subsection{Method Overview}
\bfsection{Intuition}
Our intuition is based on the observation that curvilinear structures in biological systems often exhibit tree-like morphologies, with complexity arising from two main aspects:
\begin{itemize}
\item Global structure: The overall shape and orientation of the main structure, such as the elongation and curvature of a dendrite trunk or blood vessels.
\item Local geometry: Smaller, often critical elements attached to or variations along the main structure, such as dendritic spines or vascular bifurcations.
\end{itemize}
For segmentation tasks, the global structure adds unnecessary complexity, expanding the learning space and increasing data requirements. \revised{Our approach decomposes these components by transforming the structure into standardized representations.} Such decomposition enables efficient learning through standardized \revisedsecond{cylindrical primitives} that preserve intrinsic shape information while reducing global variations.

\bfsection{Segmentation Pipeline with FFD}
We use dendritic spine segmentation as an exemplar to demonstrate the application of Frenet--Serret Frame-based Decomposition (FFD) for segmenting 3D curvilinear structures. As illustrated in Fig.~\ref{fig:teaser}, our pipeline consists of three main stages:

\begin{itemize}
\item \textit{Decomposition}: \revised{We first convert binary EM volumes to point clouds by treating each nonzero voxel in the volume as a point in $\mathbb{R}^3$ with integer coordinates.} We then perform skeletonization with topological pruning to extract the backbone (dendrite trunk) skeleton, parameterizing it as a $C^2$ continuous curve. Along this curve, we calculate Frenet--Serret Frames and reconstruct surrounding point clouds in a cylindrical coordinate system (Fig.~\ref{fig:ffd}). This forms a cylindrical primitive in $(\mathbb{R}^+ \times S^1 \times \mathbb{R})$, preserving essential local geometries.

\item \textit{Segmentation}: With its reduced learning space, the cylindrical primitive undergoes data-efficient segmentation, as well as enabling improved generalization across diverse samples. \revised{For compatibility with point-based networks that rely on Euclidean distances for operations like neighbor selection and feature aggregation, the cylindrical coordinates are transformed to Cartesian coordinates as input of the segmentation network.}

\item \textit{Inverse Decomposition}: Finally, we transform the segmented cylindrical primitive and $C^2$ curve back to the original $\mathbb{R}^3$ space, completing the process. \revisedsecond{Note that this step is optional for point-wise segmentation tasks, as the bijective property allows label predictions to be directly assigned to their corresponding points in the original space without performing the inverse transformation.}
\end{itemize}

This approach significantly boosts segmentation accuracy and generalization performance on dendritic spine segmentation task, as demonstrated in our experiments (Sec.\ref{exp:denspine}). In the following subsections, we provide the mathematical formulation of the decomposition (Sec.\ref{method:definition}), prove its properties (bijectivity and rotation-invariance, Sec.\ref{method:property}), and detail the implementation of the pipeline, including skeletonization and discrete Frenet-Serret Frame calculation (Sec.\ref{method:implementation}).

\subsection{Formulation of Frenet-Serret Frame-based Decomposition}\label{method:definition}
\begin{figure}[h]
\centering
\includegraphics[width=\linewidth]{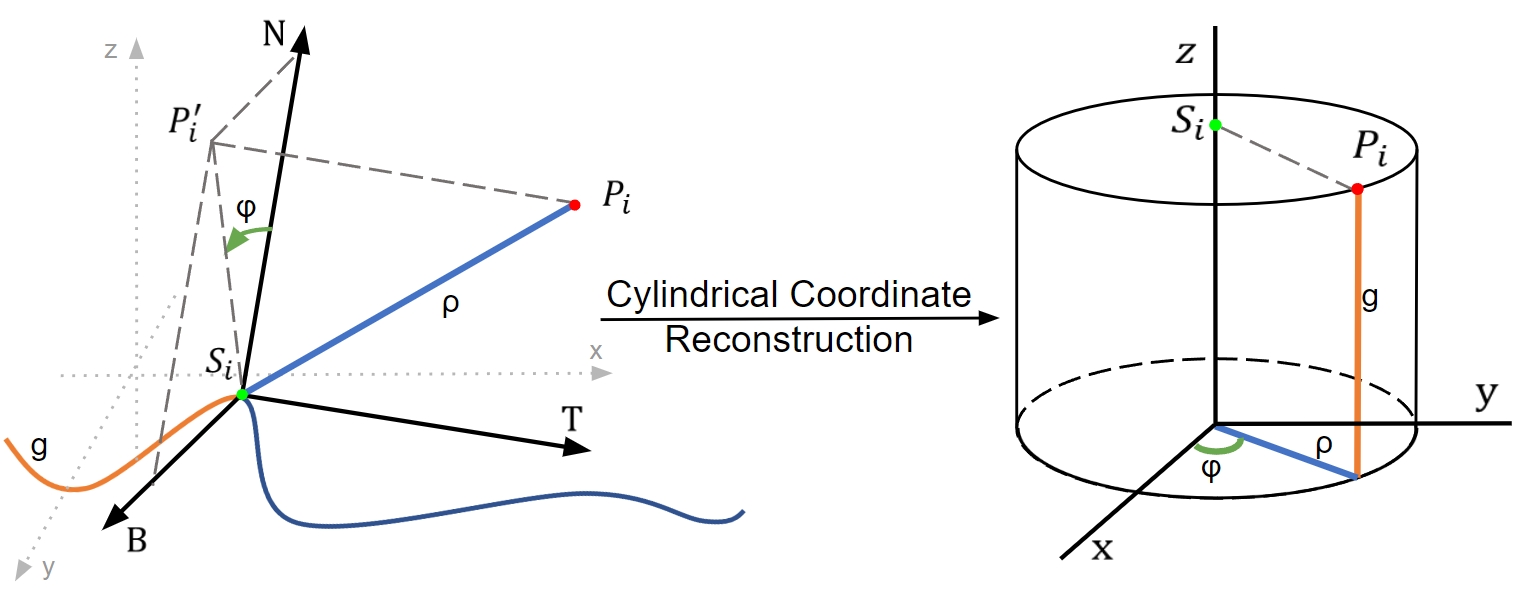}
    \caption{\textbf{Frenet-Serret Frame-based Transformation.} A key component of FFD is the transformation that maps point clouds in $\mathbb{R}^3$ to cylindrical coordinates $(\rho, \revised{\varphi}, g)$ in $(\mathbb{R}^+ \times S^1 \times \mathbb{R})$. It utilizes the Frenet-Serret Frame (T, N, B) of the curve at $S_i$, the nearest point to $P_i$. \textbf{Left}: The point $P_i$ and curve $S$ in a Cartesian coordinate system. $\rho$: distance between $S_i$ and $P_i$, \revised{$\varphi$}: angle between normal vector N and projection of $\overrightarrow{S_iP_i}$ onto NB-plane, $g$: curve arc length to $S_i$. \textbf{Right}: The reconstruction of $P_i$ in a cylindrical coordinate system.}
\label{fig:ffd}
\vspace{-1em}
\end{figure}
Denote $P = \{(x_i, y_i, z_i) \mid i = 1, \ldots, n\} \subset \mathbb{R}^3$ as a point cloud, $C$ as the space of $C^2$ curves in $\mathbb{R}^3$ that form the backbone skeleton of $P$. \revisedsecond{Note that our pipeline can accept either volumetric or point cloud input, as binary volumes can be treated as point clouds with integer coordinates. For generality, we formulate our method using point cloud notation.} We formulate the decomposition:
\begin{equation}
\mathcal{D}: P \to C \times (\mathbb{R}^+ \times S^1 \times \mathbb{R})^n, 
\end{equation}
as a composition of two mappings: $\mathcal{D} = \mathcal{S} \circ \mathcal{F}$, where:

\begin{itemize}
    \item $\mathcal{S}: P \to C$ is a skeletonization function that maps the point cloud to a $C^2$ continuous curve $S: [0,L] \to \mathbb{R}^3$, parameterized by arc length $s \in [0,L]$. 
    \item $\mathcal{F}: C \times P \to C \times (\mathbb{R}^+ \times S^1 \times \mathbb{R})^n$ is a Frenet-Serret Frame-based transformation that reconstruct the point cloud in cylindrical coordinates, defined as:
\begin{equation}
\mathcal{F}(S, P) = (S, \{(\rho_i, \revised{\varphi_i}, g_i) \mid i = 1, \ldots, n\}),
\end{equation}
\end{itemize}
where $\{(\rho_i, \revised{\varphi_i}, g_i) \mid i = 1, \ldots, n\}$ is the reconstructed point cloud in a cylindrical coordinate system.

Specifically, as depicted in Fig.~\ref{fig:ffd}, for each point $P_i$, we determine its closest point on the curve, $S_i = S(s_i)$, where $s_i = \min_{s \in [0,L]} \|P_i - S(s)\|$. Due to the continuity of $S$, the closest point is unique for almost all $P_i$\footnote{For a continuous curve, almost every point in $\mathbb{R}^3$ has a unique closest point on the curve. The set of points with multiple equally closest points (\ie, cut locus) is of measure zero and does not affect the overall transformation.}. The transformation is then defined as:
\begin{align}
& \rho_i = \|P_i - S_i\|, \quad g_i = \int_0^{s_i} \left\|\frac{dS(s)}{ds}\right\| ds = s_i \nonumber,\\
& \revised{\varphi_i} = \arctan2(\mathbf{v}_i \cdot \mathbf{b}_{s_i}, \mathbf{v}_i \cdot \mathbf{n}_{s_i})\nonumber,
\end{align}
where $\mathbf{v}_i$ represents the projection of the vector $\overrightarrow{S_iP_i}$ (denoted as $\mathbf{u}_i$) onto the normal-binormal plane, which can be calculated by $\mathbf{v}_i = A_iA_i^T\mathbf{u}_i$, where $A_i = [\mathbf{n}_{s_i}, \mathbf{b}_{s_i}]$ is a column orthogonal matrix.

\subsection{Properties of the Decomposition}\label{method:property}
\bfsection{Properties}
The Frenet--Serret Frame-based Decomposition possesses two key properties: 1)~\textit{Bijectivity}: The decomposition is invertible, allowing the cylindrical primitive and backbone curve to be transformed back to the original space without information loss. 2)~\textit{Rotation Invariance}: The decomposition is invariant to rotations of the input data, as the cylindrical primitive is constructed in a standardized coordinate system aligned with the backbone curve.

\bfsection{Benefits} These properties confer the following benefits:
1) Bijectivity enables segmentation to be performed in the simplified cylindrical space while preserving the ability to map results back to the original space accurately.
2) Rotation invariance eliminates the need for rotation augmentation and ensures consistent feature representation regardless of the input orientation.

\revised{
\bfsection{Intuitive Explanation of Properties}
To complement our formal proofs, we provide the following intuitive explanations: Our decomposition achieves bijectivity because (1) injectivity follows from the uniqueness of closest-point mapping—any two points sharing identical cylindrical coordinates would create a contradiction in the distance minimization principle; and (2) surjectivity is guaranteed as we can explicitly construct a 3D point for any valid cylindrical coordinate by positioning it at the specified distance and angle from the corresponding location on the curve. Rotation invariance occurs because our cylindrical coordinates represent invariant geometric properties: $\rho$ measures Euclidean distance, $\varphi$ captures the relative angle in the normal-binormal plane, and $g$ represents arc length—all quantities that remain unchanged under rigid rotations.
}

\bfsection{Proof}
To prove the properties of the decomposition $\mathcal{D}$, it suffices to prove the corresponding properties of $\mathcal{F}$. Given that $\mathcal{S}: P \to C$ is a fixed mapping for a given point cloud, the properties of $\mathcal{D} = \mathcal{F} \circ \mathcal{S}$ are fundamentally determined by $\mathcal{F}: C \times P \to C \times (\mathbb{R}^+ \times S^1 \times \mathbb{R})^n$. Therefore, we focus the proof on the Frenet-Serret Frame-based transformation $\mathcal{F}$. For notational convenience, we use $\mathcal{F}(P)$ to represent $\mathcal{F}(S, P)$ in our proofs, as $S$ is fixed for a given input.

\itsection{1) Bijectivity} To prove the transformation is bijective, we need to verify that it's both injective and subjective.
\begin{itemize}
\item Injectivity: Assume $P_{t1}, P_{t2} \in P$, with $\mathcal{F}(P_{t1}) = \mathcal{F}(P_{t2}) = (\rho, \revised{\varphi}, g)$. Let $S_t \in \mathbb{R}^3$ be their closest point on the skeleton $S$. If $P_{t1} \neq P_{t2}$, then $\overrightarrow{P_{t1}P_{t2}} = \delta \textit{\textbf{t}},~\delta \neq 0$, where $\textit{\textbf{t}}$ is the tangent at $S_t$. As $S$ is $C^2$ continuous, $\exists~\epsilon > 0$ sufficiently small and $S'_t \in S$ such that $\overrightarrow{S'_tS_t} = \epsilon\textit{\textbf{t}}$ and $\|\overrightarrow{S'_tP_{t1}}\|^2 = \|\overrightarrow{S_tP_{t1}}\|^2 - \epsilon^2 + o(\epsilon^2)$. Hence $d(P_{t1}, S'_t) < d(P_{t1}, S_t)$ , contradicting that $S_t$ is the closest point to $P_{t1}$ on $S$. Hence, $\forall~P_{t1}, P_{t2}\in P$ such that $\mathcal{F}(P_{t1}) = \mathcal{F}(P_{t2})$, we have $P_{t1} = P_{t2}$, \ie, the transformation is injective.

\item Surjectivity: As $S$ is $C^2$ continuous, $\forall s_1, s_2 \in [0,L]$ $(s_1 \neq s_2)$, we have $\|S_{s_1} - S_{s_2}\| > 0$. Hence, $\forall Y_t = (\rho_t, \revised{\varphi_t}, g_t) \in \mathbb{R}^+ \times S^1 \times \mathbb{R}$, $S_{s_t} \in S$ can be uniquely determined by $g_t = \int_0^{s_t} \|\frac{dS(s)}{ds}\| ds$. Denote the Frenet-Serret Frame at $S_{s_t}$ as $(\textit{\textbf{t}}_{s_t},\textit{\textbf{n}}_{s_t},\textit{\textbf{b}}_{s_t})$. $\exists~\delta\in(0,\rho_t)$, we have $P_t \in \mathbb{R}^3$ as $\overrightarrow{S_{s_t}P_t} = \delta(\sin\revised{\varphi_t}\textit{\textbf{b}}_{s_t} +\cos\revised{\varphi_t}\textit{\textbf{n}}_{s_t})+\sqrt{{\rho_{t}^{2}}-{\delta^{2}}}\textit{\textbf{t}}_{s_t}$, such that $\mathcal{F}(P_t) = S_t$. Hence, $\forall Y_t \in \mathbb{R}^+ \times S^1 \times \mathbb{R}$, $\exists P_t \in \mathbb{R}^3$ such that $\mathcal{F}(P_t) = Y_t$, \ie, the transformation is surjective.
\end{itemize}

\itsection{2) Rotation Invariance}
We prove the rotation invariance of $\mathcal{F}$ by showing $\mathcal{F}(R(P_t)) = \mathcal{F}(P_t)$ for any $P_t \in \mathbb{R}^3$ and $R \in SO(3)$. 

Let $S_{s_t}$ be the closest point on $S$ to $P_t$, $(\mathbf{t}(s_t),\mathbf{n}(s_t),\mathbf{b}(s_t))$ the Frenet-Serret Frame of $S$ at $S_{s_t}$, $\mathbf{u}_t = \overrightarrow{S_{s_t}P_t}$, $A_t = [\mathbf{n}(s_t), \mathbf{b}(s_t)]$, and $\mathbf{v}_t = A_tA_t^T\mathbf{u}_t$. Under rotation $R$, the Frenet-Serret Frame rotates accordingly: $(\mathbf{t}'(s_t),\mathbf{n}'(s_t),\mathbf{b}'(s_t)) = (R\mathbf{t}(s_t),R\mathbf{n}(s_t),R\mathbf{b}(s_t))$, and $\mathbf{u}'_t = R(\mathbf{u}_t)$, $A'_t = R(A_t)$. Hence, $\mathbf{v}'_t = R(\mathbf{v}_t)$. Crucially:
\begin{align*}
(a)\quad &\rho'_t = \|R(P_t) - R(S_{s_t})\| = \|P_t - S_{s_t}\| = \rho_t \\
(b)\quad &g'_t = g_t \\
(c)\quad &\revised{\varphi'_t} = \arctan2(R(\mathbf{v}_t) \cdot R(\mathbf{b}(s_t)), R(\mathbf{v}_t) \cdot R(\mathbf{n}(s_t))) \\
&\phantom{\revised{\varphi'_t}} = \arctan2(\mathbf{v}_t \cdot \mathbf{b}(s_t), \mathbf{v}_t \cdot \mathbf{n}(s_t)) = \revised{\varphi_t}
\end{align*}
(a), (b), and (c) hold because rotation preserves distances, arclength, and dot products, respectively. Thus, $\mathcal{F}(R(P_t)) = (\rho'_t, \revised{\varphi'_t}, g'_t) = (\rho_t, \revised{\varphi_t}, g_t) = \mathcal{F}(P_t)$, ensuring consistent transformation regardless of orientation.

\subsection{Implementation \revised{of the Pipeline} }\label{method:implementation}
\bfsection{Backbone Skeletonization}\label{method:skeletonization}
We first apply the \textit{Tree-structure Extraction Algorithm for Accurate and Robust Skeletons} (TEASAR)~\cite{TEASAR} to extract the initial skeleton from the input structure. TEASAR begins with a raster scan to locate an arbitrary foreground point, identifying its furthest point as the root. It then implements Euclidean distance transform to define a penalty field~\cite{Bitter2001PenalizedDistanceVS}, guiding the skeleton through the target's center. Dijkstra's algorithm is applied to find the path from the root to the most geodesically distant point, forming a skeleton branch. Visited regions are marked by expanding a circumscribing cube around the path vertices. This process repeats until all points are traversed. Finally, the resultant skeleton is smoothed and upsampled via linear interpolation for density assurance.  

\begin{figure}[t]
\centering
\includegraphics[width=\linewidth]{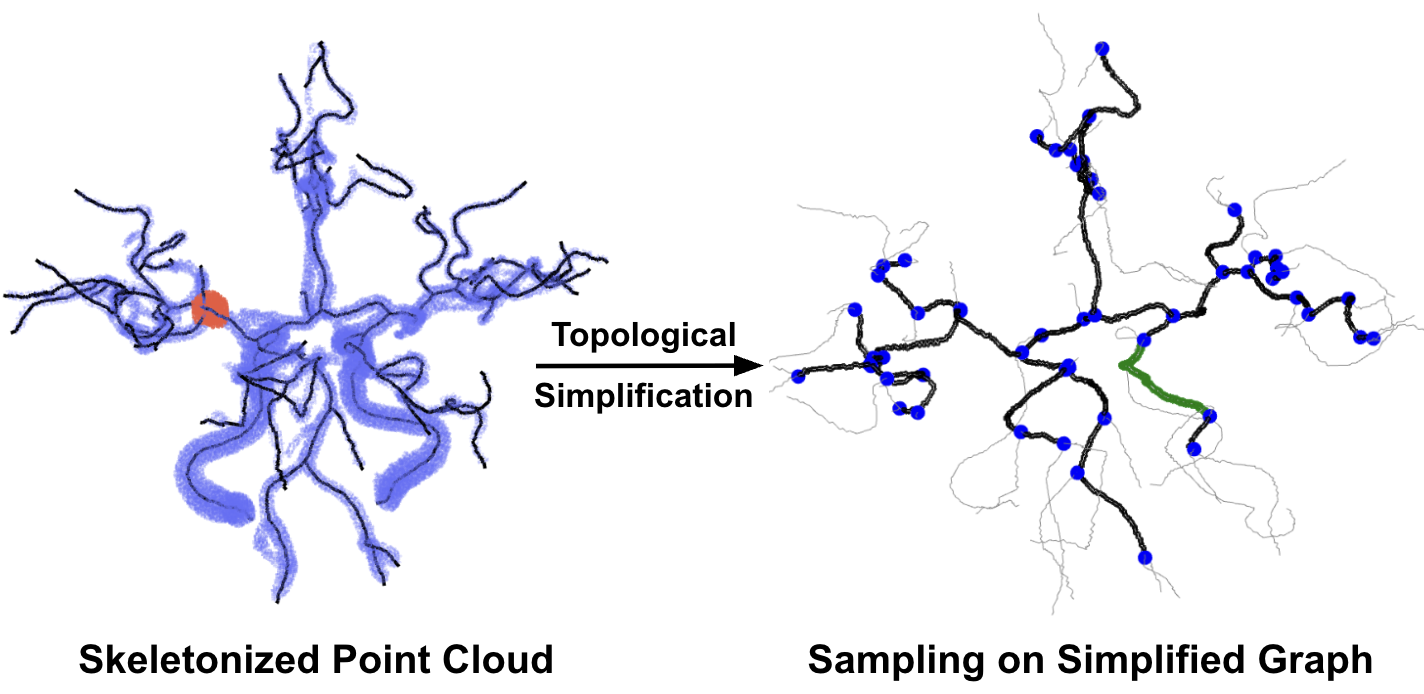}


\caption{\revisedsecond{\textbf{Demonstration of Sampling Process on IntrA.} \textbf{Left:} Artery point cloud showing vessels (blue) and aneurysm (red), with the extracted skeleton (black) forming a tree-structure graph. \textbf{Right:} Topological simplification identifies all junction nodes (blue dots), and the paths connecting neighboring junction nodes collectively represent the main structure. Each path between neighboring junction nodes represents a distinct segment for processing (e.g., the green path). If a path is shorter than a predefined threshold, we extend it by including adjacent paths to ensure sufficient context.
}}

\vspace{-1em}
\label{fig:skeleton_sample}
\end{figure}


\begin{figure*}[tb]
\centering
\includegraphics[width=\linewidth]{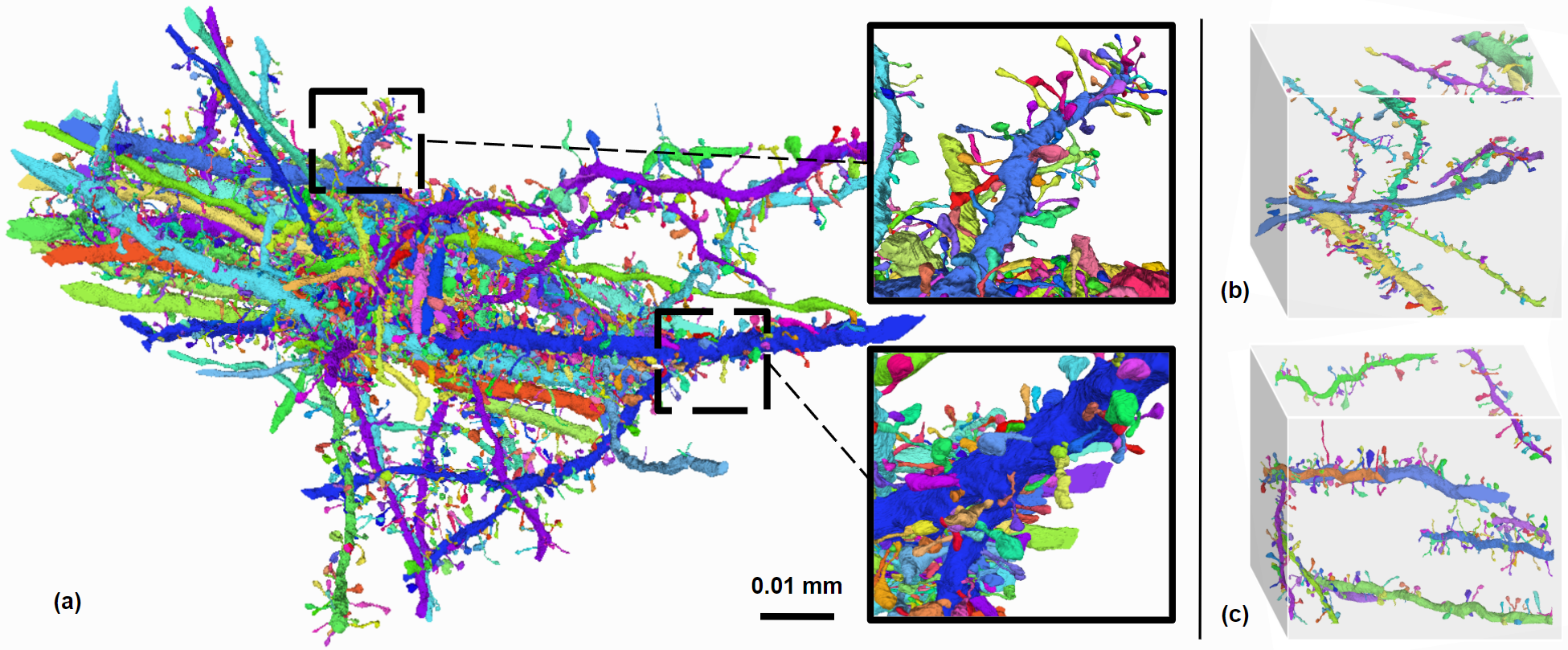}
\caption{\textbf{DenSpineEM Dataset.} DenSpineEM contains 3 subsets: (a) M50: 50 mouse somatosensory cortex dendrites, (b) H10: 10 human visual cortex dendrites, (c) M10: 10 mouse visual cortex dendrites.} \label{fig:dataset_denspineem}
\vspace{-1em}
\end{figure*}

\revised{\bfsection{Branch Extraction} 
Our branch extraction begins with applying the TEASAR algorithm\revisedsecond{\footnote{\url{https://github.com/seung-lab/kimimaro}}} to skeletonize the complete structure, \revisedsecond{which directly outputs a tree graph representation (vertices connected by edges). Before traversal, we manually inspect the TEASAR result and prune obvious artifacts such as redundant branches or spurious loops.} We \revisedsecond{then} traverse this graph and identify leaf nodes (degree 1) as branch endpoints and junction nodes (degree $> 2$) as intersection points. This generates multiple paths connecting these key nodes. \revisedsecond{We perform topological simplification by pruning all leaf branches, keeping only the paths connecting junction nodes as the main structure.} We sample \revisedsecond{each edge of the simplified graph} as distinct segments\revisedsecond{, ensuring all edges are processed. For segments shorter than the threshold length, we extend them by including adjacent paths at junction nodes}. \revisedsecond{Fig.~\ref{fig:skeleton_sample} illustrates this topological simplification and sampling process on a sample from the IntrA dataset.} The sampling length is dataset-specific: $10,000~nm$ for DenSpineEM to ensure adequate curvature and spine presence, and $100~mm$ for IntrA based on the aneurysm size distribution in \cite{Yang2020ATS}. \revisedsecond{To obtain point cloud segments, we first assign all points to their closest skeleton vertices based on Euclidean distance in Cartesian space. For each sampled segment, we then extract the subset of points assigned to vertices within that segment, including both the main structure edge and any pruned branches connected to it.} This approach effectively decomposes complex structures into manageable segments for subsequent analysis.
}




\revised{\bfsection{Skeletonization Algorithm Choice}} Our pipeline is flexible and can be adapted to process both volumetric and point cloud inputs. \revised{In this study, TEASAR is applied to both DenSpineEM and IntrA experiments, while for the CurviSeg experiment, the skeleton is provided in the synthetic data generation procedure.} Alternatively, we refer to L1-medial skeletonization~\cite{Huang2013L1medialSO} as a robust approach for small-scale point clouds.


\bfsection{Discrete Frenet-Serret Frame Computation}
We compute Frenet-Serret Frames along the curve to characterize local geometry, addressing both curved and straight segments. For curved segments, we apply standard Frenet-Serret formulas as defined in Eq.~\ref{eq:tnb}. To enhance numerical stability, we employ a curvature threshold $\epsilon = 1e-8$, identifying near-straight segments where Frenet-Serret Frames become ill-defined. Our piecewise interpolation scheme handles straight segments effectively. Between curved parts, we linearly interpolate the normal vector, while at curve extremities, we propagate the normal from the nearest curved segment. \revised{For globally straight curves, we define a constant normal vector along the curve by first selecting an endpoint and its adjacent point to compute the tangent direction. We then define an arbitrary vector that is not parallel to the tangent, construct a normal vector by cross-product to ensure orthogonality, and apply it consistently along the entire curve.
}

To ensure frame orthonormality and further improve numerical stability, we apply Gram-Schmidt orthogonalization\revised{\cite{Chrzeszczyk2011MatrixC}}. Our Frenet-Serret Frame computation method is provided as a Python package\footnote{\url{https://pypi.org/project/discrete-frenet-solver}}, facilitating seamless integration into various geometric analysis and computational applications.

\revised{
\bfsection{Cylindrical Primitive for Segmentation}
For each cropped branch, we first compute the Frenet–Serret frame along its skeleton. Following the procedure described in Section~\ref{method:definition}, the corresponding point cloud is then transformed into a standardized cylindrical representation.

As point-based networks rely on Euclidean distances for operations such as neighbor selection and feature aggregation, it is essential to convert the cylindrical coordinates back to Cartesian coordinates before segmentation. Specifically, given the cylindrical coordinates \((\rho, \varphi, g)\) for each point, the Cartesian coordinates \((x,y,z)\) are computed as:
\begin{equation}
x = \rho \cos \varphi,\quad y = \rho \sin \varphi,\quad z = g.
\end{equation}
This conversion ensures that all distance-based computations remain valid while preserving the advantages of the cylindrical representation.

\bfsection{Inverse Decomposition}
The inverse decomposition $\mathcal{D}^{-1}: C \times (\mathbb{R}^+ \times S^1 \times \mathbb{R})^n \to \mathbb{Z}^{3 \times n}$ reconstructs the original integer point cloud from the backbone curve and cylindrical coordinates. Given the curve $S: [0,L] \to \mathbb{R}^3$ and the cylindrical coordinates $\{(\rho_i, \revised{\varphi_i}, g_i) \mid i = 1, \ldots, n\}$, the inverse transformation is defined component-wise as:
\begin{equation}
P_i = \lfloor S(g_i) + \rho_i (\cos(\revised{\varphi_i}) \mathbf{n}_{g_i} + \sin(\revised{\varphi_i}) \mathbf{b}_{g_i}) + 0.5 \rfloor
\end{equation}\label{eq:inverse_decompose}
where $S(g_i)$ is the point on the curve at arc length $g_i$, $\mathbf{n}_{g_i}$ and $\mathbf{b}_{g_i}$ are the normal and binormal vectors of the Frenet-Serret frame at that point, and $\lfloor \cdot + 0.5 \rfloor$ denotes component-wise rounding to the nearest integer. This rounding operation compensates for the loss of tangential information when projecting points onto the normal-binormal plane during the forward transformation. As the sampling density of $S$ approaches infinity, Eq.~\ref{eq:inverse_decompose} can recover the original integer point cloud with arbitrary precision.

In practical implementations, to reconstruct the original point cloud, the tangential component $h_i = \mathbf{u}_i \cdot \mathbf{t}_{s_i}$ can be stored for reconstruction via $P_i = S(g_i) + \rho_i (\cos(\revised{\varphi_i}) \mathbf{n}_{g_i} + \sin(\revised{\varphi_i}) \mathbf{b}_{g_i}) + h_i \mathbf{t}_{g_i}$. This recovers the full displacement vector by adding back the component along the tangent direction, which was omitted in the cylindrical representation. 

For tasks requiring only point-wise predictions rather than complete geometric reconstruction, we leverage the bijective property to eliminate the need for performing the inverse transformation, as labels predicted in the cylindrical space can be directly assigned to their corresponding points in the original point cloud, substantially reducing computational overhead.

}

\section{Datasets}
\subsection{\revised{CurviSeg} Dataset}
We introduce the \revised{CurviSeg} dataset and make use of it for the first experiments in this paper. 
\revised{CurviSeg} is defined as a synthetic dataset of 3D curvilinear structures with additional spherical objects for point cloud segmentation tasks. The curvilinear structures were generated using cubic B-spline interpolation of $n$ randomly generated control points, where $n \sim \mathcal{U}\{5, 10\}$. The control points $\mathbf{p}_i \in \mathbb{R}^3$ were generated as:
\begin{equation}
    \mathbf{p}_i = s \cdot \mathbf{r}_i, \quad i = 1, \ldots, n
\end{equation}
where $\mathbf{r}_i \sim \mathcal{N}(\mathbf{0}, \mathbf{I}_3)$ are random vectors sampled from a standard 3D normal distribution, and $s \sim \mathcal{U}(1, 3)$ is a uniform random scaling factor. The B-spline curve $\mathbf{C}(t)$ was then defined as:
\begin{equation}
    \mathbf{C}(t) = \sum_{i=0}^{n-1} N_{i,3}(t) \mathbf{p}_i, \quad t \in [0, 1]
\end{equation}
where $N_{i,3}(t)$ are cubic B-spline basis functions. This curve was evaluated at 500 equidistant points $\{t_j\}_{j=1}^{500}$ to form the skeleton. Points were distributed along this skeleton using a cylindrical coordinate system. For each skeleton point $\mathbf{C}(t_j)$, we generated a set of points $\mathbf{x}_{j,k}$ as:
\begin{equation}
    \mathbf{x}_{j,k} = \mathbf{C}(t_j) + r \cos(\theta) \mathbf{n}_j + r \sin(\theta) \mathbf{b}_j
\end{equation}
where $r \sim \mathcal{U}(0, r_s)$, $r_s \sim \mathcal{U}(0.3, 0.7)$ is the slice radius, $\theta \sim \mathcal{U}(0, 2\pi)$, and $\mathbf{n}_j$ and $\mathbf{b}_j$ are the normal and binormal vectors at $\mathbf{C}(t_j)$, respectively. 
We added $m \sim \mathcal{U}\{1, 2, 3\}$ spherical objects to each structure. Each sphere, centered at $\mathbf{c}_l$, was placed tangent to a random point $\mathbf{x}_{j,k}$ on the main structure:
\begin{equation}
    \mathbf{c}_l = \mathbf{x}_{j,k} + (r_s + r_b) \frac{\mathbf{x}_{j,k} - \mathbf{C}(t_j)}{\|\mathbf{x}_{j,k} - \mathbf{C}(t_j)\|}
\end{equation}
where $r_b = kr_s$, $k \sim \mathcal{U}(1, 2)$. Points within each sphere were generated as:
\begin{equation}
    \mathbf{y}_l = \mathbf{c}_l + r_b \cdot \mathbf{u}, \quad \mathbf{u} \sim \mathcal{U}(\mathbb{S}^2)
\end{equation}
where $\mathbb{S}^2$ is the unit 2-sphere. The point density was kept consistent between the main structure and the spheres, calculated based on the total volume and target point count. Each point was labeled as either part of the main structure (0) or a sphere (1), forming a binary segmentation problem. 

\revised{CurviSeg} comprises $2500$ samples in total, where each sample contains $4096$ points. The dataset is split into $80\%$ training, $10\%$ validation, and $10\%$ testing sets.

\subsection{DenSpineEM \revised{Benchmark}}
We curate a large-scale 3D dendritic spine segmentation benchmark, \textit{DenSpineEM}, with saturated manual annotation of three EM image volumes (Fig.~\ref{fig:dataset_denspineem}). 
In total, DenSpineEM contains \revised{4,476} spine instances from \revised{70} fully segmented dendrites (Tab.~\ref{tab:dataset}).
In comparison, existing dendrite spine segmentation datasets are either constructed by heuristic spine extraction methods~\cite{wildenberg2021large, dorkenwald2019binary} or lack of thorough annotation~\cite{ofer2021ultrastructural}.




\begin{table*}[t]
\caption{\label{tab:dataset}\textbf{Overview of DenSpineEM Dataset.} We build upon 3 EM volumes with instance segmentation and annotate spine segmentation for 70 dendrites.}
\centering
\setlength{\tabcolsep}{0pt}
\begin{tabular*}{0.8\textwidth}{@{\extracolsep{\fill}}lcccr}
\hline
{\bf Name} & {\bf Tissue} & {\bf Size ($\mu m^3$)}& {\bf \#Dendrites} & {\bf \#Spines}\\
\hline
DenSpine-M50 & Mouse Somatosensory Cortex~\cite{Kasthuri2015} &50$\times$50$\times$30 &50 & 3,827\\
DenSpine-M10 & Mouse Visual Cortex~\cite{wei2021axonem}&30$\times$30$\times$30 & 10 &335 \\
DenSpine-H10 & Human Frontal Lobe~\cite{wei2021axonem}&30$\times$30$\times$30& 10 &\revised{314}\\
\hline
\end{tabular*}
\end{table*}

\bfsection{Dataset Construction}
We leverage three public EM image volumes with dense dendrite segmentation to construct the DenSpineEM dataset: one \revised{$50\times 50\times 50 ~ \mu m^3$} volume from the mouse somatosensory cortex~\cite{Kasthuri2015}, two \revised{$30\times 30\times 30 ~ \mu m^3$} volumes from the mouse visual cortex and the human frontal lobe respectively~\cite{wei2021axonem} (Tab.~\ref{tab:dataset}).
We refer readers to the references for dataset details.

\itsection{DenSpine-M50} We first curate DenSpine-M50 from~\cite{Kasthuri2015} as our main dataset due its existing segmented dendrites (100+) and spines (4,000+) which are analyzed in~\cite{ofer2021ultrastructural}. 
However, the spine segmentation on most dendrites is not thorough, making it difficult to train models for practical use due to false negative errors. 
We pick the 50 largest dendrites from the existing annotation and manually proofread all spine instance segmentation.
In the end, we obtain 3,827 spine instances. 

\itsection{DenSpine-\{M10, H10\}} To evaluate the generalization performance of the model trained on DenSpine-M50 across regions and species, we build two additional datasets from AxonEM image volumes~\cite{wei2021axonem}: DenSpine-M10 from another brain region in the mouse and DenSpine-H10 from the human.
Although the AxonEM dataset only provides proofread axon segmentation, we are thankful to receive saturated segmentation results for both volumes from the authors. 
For each of the two volumes, we first pick 10 dendrites with various dendrite types and branch thicknesses and proofread their segmentation results. Then, we go through these dendrites and annotate the spine instance segmentation.

\bfsection{Annotation Protocol}
To generate high-quality ground truth annotations, we segment spines manually with the VAST software~\cite{berger2018vast} to avoid introducing bias from automatic methods. 
To detect errors, we use the neuroglancer software~\cite{ng} to generate and visualize 3D meshes of the segmentation of dendrites and spines. Four neuroscience experts were recruited to proofread and double-confirm the annotation results for spine instance segmentation.


\begin{table}[t]
\centering
\small
\caption{\textbf{FFD Validation on CurviSeg.} We evaluate the segmentation performance, data efficiency, rotation invariance, and computation speed of three models with and without FFD. }
\label{tab:fft_impact}
\resizebox{.5\textwidth}{!}{%
\begin{tabular}{l|c c c|c c}
\toprule
\multirow{2}{*}{\textbf{Method}} & \multicolumn{3}{c|}{\textbf{Segmentation Performance (DSC \%)}} & \multicolumn{2}{c}{\textbf{Computation Speed}} \\
& Full Data & 25\% Data & Test-time Rot. & Train (s/epoch) & Inf. (ms/sample) \\
\midrule
PointCNN\cite{Li2018PointCNNCO} & 92.40 & 84.92 & 91.32 & 210.00 & 119.59 \\
w. FFD &  95.42 & 94.77 & 95.60 & 215.40 & 124.78  \\
\midrule
PointNet++\cite{qi2017pointnet++} & 87.99 & 56.91 & 85.87 & 75.81 & 32.34 \\
w. FFD & 95.17 & 94.33 & 95.18 & 82.70 & 38.05  \\
\midrule
DGCNN\cite{Wang2018DynamicGC} & 88.95 & 84.32 & 86.41 & 114.11 & 58.87 \\
w. FFD & 95.76 & 95.63 & 95.76 & 122.33 & 63.06  \\
\bottomrule

\end{tabular}%
}
\end{table}


\subsection{IntrA Dataset}
\revised{\bfsection{Dataset Overview}
The IntrA dataset~\cite{Yang2020IntrA3I} is a publicly available benchmark for intracranial aneurysm segmentation. It provides two components: (1) 103 complete 3D artery models with aneurysms, and (2) 1909 blood vessel segments (1694 healthy segments and 215 aneurysm segments) for classification and segmentation benchmarks.

\bfsection{Experimental Settings}
Most studies focus on segmenting aneurysms from vessel segments, which are short and less complex, resulting in high performance (e.g., 89.71\% DSC on aneurysm segmentation~\cite{Yu20213DMP}). In contrast, we target aneurysm segmentation on full artery models—a more realistic, clinically relevant, and challenging task. To our knowledge, the state-of-the-art model on this task achieves only 71.79\% DSC on aneurysm segmentation~\cite{Yang2020ATS}.

\bfsection{Evaluation on Full Artery Models}}
As additional evaluation, we use the entire artery subset of the IntrA dataset~\cite{Yang2020ATS} (instead of the commonly used segment subset~\cite{Yang2020IntrA3I}). This subset consists of 103 3D TOF-MRA images containing 114 aneurysms. The data are provided as surface models in Wavefront OBJ files, derived from original volumetric images (512 × 512 × 300, 0.496 mm slice thickness). Using full artery models presents a more challenging and realistic scenario for aneurysm segmentation. The dataset excludes aneurysms smaller than 3.00 mm, with sizes ranging from 3.48 to 18.66 mm (Mean: 7.49 mm, SD: 2.72 mm). Most aneurysms are saccular, with one fusiform aneurysm included.


\begin{figure}[t]
\centering
\includegraphics[width=\linewidth]{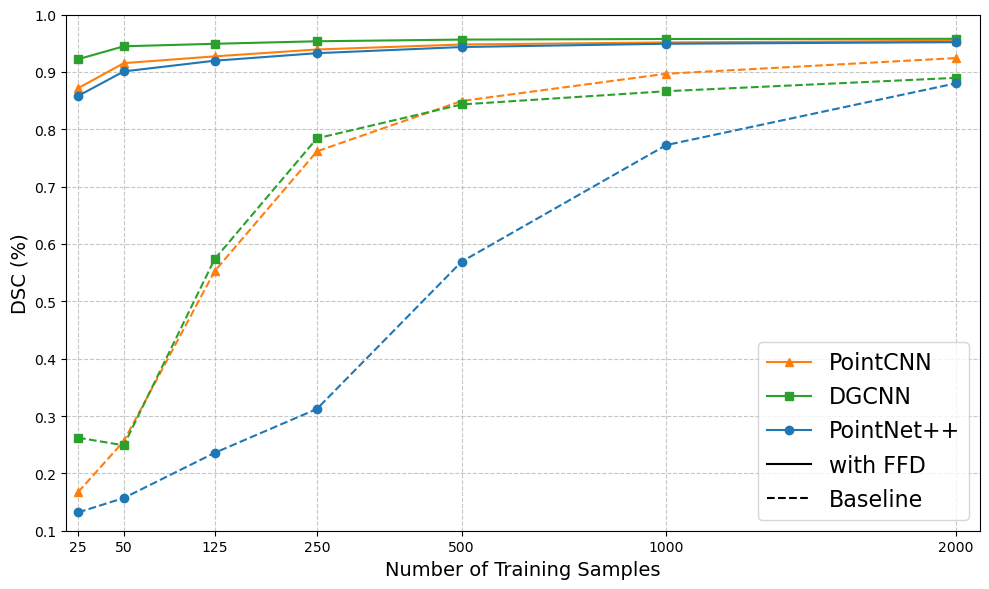}
\caption{\revised{\textbf{Data Efficiency Plot of FFD.} We compare models trained on varying scales of data from CurviSeg dataset.}}
\vspace{-1em}
\label{fig:data_efficiency}
\end{figure}

\revised{
\begin{table*}[t]
\centering
\caption{\textbf{Segmentation Results on DenSpineEM.} The results are calculated by the mean value of each fold. 95\% confidence intervals are given in parentheses.}
\resizebox{\textwidth}{!}{%
\revised{
\begin{tabular}{l|c|cc|cc|c|c}
\hline
\multirow{2}{*}{Method} & \multirow{2}{*}{Subset} & \multicolumn{2}{c|}{IoU (\%)} & \multicolumn{2}{c|}{DSC (\%)} & \multirow{2}{*}{Spine Accuracy (\%)} & \multirow{2}{*}{Spine Recall (\%)} \\
\cline{3-6}
 &  & Spine & Trunk & Spine & Trunk &  &  \\
\hline
\multirow{3}{*}{PointNet++\cite{qi2017pointnet++}} & M50 & 69.51 (66.76 - 72.26) & 94.84 (92.89 - 96.80) & 80.04 (77.53 - 82.54) & 97.24 (96.08 - 98.40) & 87.03 (85.37 - 88.69) & 86.44 (85.13 - 87.75) \\
 & M10 & 73.29 (71.39 - 75.18) & 89.48 (88.65 - 90.31) & 84.28 (82.99 - 85.57) & 94.33 (93.84 - 94.81) & 78.66 (75.87 - 81.45) & 80.58 (78.26 - 82.91) \\
 & H10 & 60.56 (58.07 - 63.06) & 89.85 (88.80 - 90.90) & 74.19 (72.03 - 76.36) & 94.54 (93.95 - 95.13) & 66.73 (62.03 - 71.43) & 74.17 (71.29 - 77.05) \\
\hline
\multirow{3}{*}{PointNet++ w. FFD} & M50 & 82.60 (79.02 - 86.18) & 98.08 (97.78 - 98.38) & 89.28 (86.81 - 91.76) & 99.03 (98.87 - 99.18) & 89.92 (88.06 - 91.78) & 88.61 (87.25 - 89.97) \\
 & M10 & 87.40 (86.63 - 88.16) & 95.66 (95.40 - 95.92) & 93.19 (92.74 - 93.65) & 97.77 (97.63 - 97.91) & 86.56 (84.52 - 88.60) & 85.46 (83.62 - 87.29) \\
 & H10 & 68.29 (66.47 - 70.11) & 92.86 (92.24 - 93.49) & 79.27 (77.51 - 81.04) & 96.26 (95.91 - 96.61) & 78.50 (76.40 - 80.61) & 80.95 (79.04 - 82.86) \\
\hline
\multirow{3}{*}{RandLA-Net\cite{Hu2019RandLANetES}} & M50 & 14.88 (12.67 - 17.10) & 63.97 (54.85 - 73.09) & 24.63 (21.48 - 27.78) & 76.60 (69.24 - 83.96) & 43.04 (28.21 - 57.88) & 48.40 (33.96 - 62.83) \\
 & M10 & 24.33 (21.42 - 27.24) & 48.21 (42.13 - 54.29) & 37.94 (34.27 - 41.62) & 64.01 (58.15 - 69.86) & 48.85 (37.76 - 59.95) & 53.96 (43.28 - 64.64) \\
 & H10 & 22.02 (19.77 - 24.26) & 55.97 (49.24 - 62.71) & 34.49 (31.64 - 37.35) & 70.78 (65.25 - 76.31) & 49.07 (34.25 - 63.90) & 53.82 (39.51 - 68.13) \\
\hline
\multirow{3}{*}{RandLA-Net w. FFD} & M50 & 37.10 (21.49 - 52.72) & 86.65 (82.38 - 90.91) & 49.38 (33.77 - 65.00) & 92.52 (90.00 - 95.05) & 48.67 (30.08 - 67.25) & 55.24 (39.40 - 71.08) \\
 & M10 & 44.06 (32.39 - 55.74) & 78.57 (74.00 - 83.14) & 58.84 (47.12 - 70.56) & 87.62 (84.66 - 90.57) & 37.13 (24.83 - 49.44) & 48.47 (37.59 - 59.36) \\
 & H10 & 37.45 (26.63 - 48.26) & 79.00 (74.57 - 83.43) & 51.68 (40.62 - 62.73) & 87.96 (85.22 - 90.70) & 42.11 (25.31 - 58.90) & 52.99 (38.74 - 67.24) \\
\hline
\multirow{3}{*}{PointTransformer\cite{Wu2023PointTV}} & M50 & 88.07 (85.95 - 90.19) & 98.29 (98.04 - 98.53) & 92.61 (90.82 - 94.39) & 99.12 (98.99 - 99.25) & \textbf{95.94 (95.19 - 96.68)} & \textbf{94.64 (93.88 - 95.40)} \\
 & M10 & 83.77 (82.62 - 84.93) & 93.37 (92.92 - 93.83) & 90.80 (90.11 - 91.49) & 96.50 (96.25 - 96.75) & 91.21 (90.28 - 92.14) & 90.20 (89.43 - 90.97) \\
 & H10 & 73.77 (72.43 - 75.11) & 92.82 (92.37 - 93.27) & 83.95 (82.98 - 84.93) & 96.18 (95.91 - 96.44) & 82.55 (81.40 - 83.71) & 84.78 (83.87 - 85.68) \\
\hline
\multirow{3}{*}{PointTransformer w. FFD} & M50 & \textbf{90.74 (88.55 - 92.93)} & \textbf{99.00 (98.75 - 99.25)} & \textbf{94.43 (92.58 - 96.28)} & \textbf{99.49 (99.36 - 99.62)} & 95.45 (94.24 - 96.66) & 93.82 (92.46 - 95.18) \\
 & M10 & \textbf{91.72 (91.53 - 91.92)} & \textbf{97.21 (97.13 - 97.28)} & \textbf{95.61 (95.50 - 95.72)} & \textbf{98.58 (98.54 - 98.61)} & \textbf{95.86 (95.55 - 96.17)} & \textbf{93.03 (92.69 - 93.37)} \\
 & H10 & \textbf{77.57 (76.42 - 78.71)} & \textbf{95.75 (95.59 - 95.92)} & \textbf{86.63 (85.78 - 87.49)} & \textbf{97.82 (97.74 - 97.91)} & \textbf{83.68 (82.38 - 84.98)} & \textbf{84.82 (83.79 - 85.86)} \\
\bottomrule
\end{tabular}%
}
}
\label{tab:model_comparison}
\end{table*}
}

\begin{figure*}[tb]
\centering
\includegraphics[width=\linewidth]{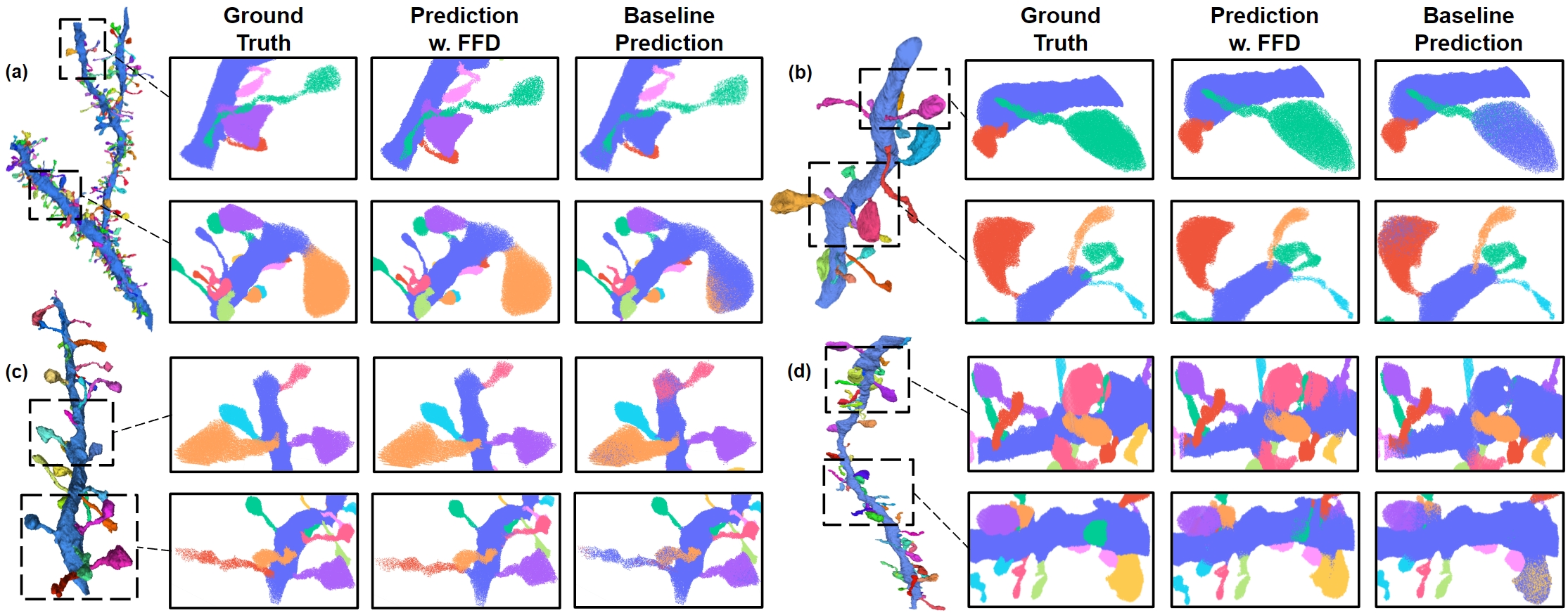}
\caption{\textbf{Segmentation Results of PointTransformer on DenSpineEM.} \revised{Models trained on M50 (Mouse somatosensory cortex, 5-fold cross-validation) and evaluated on M10 (Mouse visual cortex) and H10 (Human frontal lobe) datasets.
\textbf{(a$\sim$b) M50 result:} Both methods perform well, though baseline shows false negatives on large spines while FFD prediction provides cleaner segmentation. 
\textbf{(c) M10 result:} FFD prediction shows some false negatives at spine-trunk junctions, while baseline produces more false positives (top) and false negatives (bottom). 
\textbf{(d) H10 result:} With longer, denser spines, both methods show performance degradation, but baseline exhibits substantially more false negatives. \revisedsecond{Instance labels are assigned by overlaying ground truth; unmatched regions inherit the label of the nearest assigned prediction point.}
} }\label{fig:res_vis_denspine}

\end{figure*}
\section{Experiments and Results}

\subsection{Property Validation with CurviSeg Dataset}
We validate FFD on the CurviSeg toyset with three point-based models, using a batch size of 8 on a single A100 GPU, and assess segmentation performance with Dice.

\bfsection{Segmentation Performance}
As shown in Tab.~\ref{tab:fft_impact}, FFD consistently improved segmentation performance across all models, with 3.01\%$\sim$7.18\% increase in DSC.

\bfsection{Data Efficiency} 
We compared models trained on varying data scales, from 25 to 2000 samples. As shown in Fig.~\ref{fig:data_efficiency}, models with FFD maintain high, stable performance across all data regimes, while baseline models experience sharp performance declines as data reduces. Notably, models with FFD trained on just 25\% of the data (500 samples) perform similarly to those trained on the full dataset.

\bfsection{Rotation Invariance}
We began by applying random SE(3) augmentation\footnote{\revised{Here, SE(3) denotes the group of 3D rigid body transformations, comprising all combinations of rotation and translation that preserve the object's shape and size in 3D space.}} during test time. As shown in Tab.~\ref{tab:fft_impact}, with FFD, segmentation performance remained unchanged under rotations while non-FFD models experienced slight drops of 1.08\%$\sim$2.54\%. We further conducted a numerical analysis with 1000 SE(3)-augmented samples, comparing the representations $\mathcal{F}(P)$ and $\mathcal{F}(P_R)$. The average point-wise L2 distance was $\epsilon = (6.28 \times 10^{-26} \pm 9.13 \times 10^{-25})$, with a maximum distance of $1.85 \times 10^{-23}$, confirming the rotation invariance.

\bfsection{Computational Efficiency}
The application of FFD introduced a marginal increase in computational cost. For the training set of 2000 samples, FFD resulted in approximately $5.40s\sim8.22s$ increase in training time per epoch and $4.19ms\sim5.71ms$ increase in inference time per sample, but this trade-off was minor compared to the notable improvements in segmentation performance.

\bfsection{Bijectivity}
To empirically verify bijectivity, we randomly selected 1000 samples from CurviSeg and applied FFD, $\mathcal{D}: P \to C \times (\mathbb{R}^+ \times S^1 \times \mathbb{R})^n$, followed by its inverse, $\mathcal{D}^{-1}: C \times (\mathbb{R}^+ \times S^1 \times \mathbb{R})^n \to P'$. The average point-wise L2 distance between $P$ and $P'$ was $\epsilon = (8.98 \pm 7.21) \times 10^{-31}$, with a maximum error of $1.02 \times 10^{-29}$. These results confirm FFD’s bijectivity within numerical precision limits, demonstrating consistently low reconstruction errors across all samples.

\subsection{Benchmark on Dendritic Spine Segmentation}\label{exp:denspine}

\bfsection{Experiment Setup}
We employ 5-fold cross-validation to train models on the M50 subset, with the M10 and H10 subsets used as test sets to evaluate cross-region and cross-species generalization, respectively. Given the extreme density of input dendrite volumes—ranging from $5.59\times10^6$ to $3.51\times10^8$ voxels, with an average of $4.82\times10^7$ and the sparse spine volume (0.077\% to 6.99\% of the dendrite), voxel-based models such as nnUNet struggle with the imbalance and requires prohibitively high memory. To address the density issue, we crop dendrites along trunk skeletons and convert them into point clouds as individual samples (Sec.~\ref{method:skeletonization}). During training, \revisedsecond{we use uniform random sampling to select 30,000 points in Cartesian coordinates from each transformed point cloud; during inference, we perform repeated sampling of non-overlapping points with aggregation} to ensure full point cloud coverage. We choose 30,000 as the sampling scale as it's sufficient to preserve spine geometry and shapes, whereas fewer points risk losing critical information.

\bfsection{Model Choice}
Although 30,000 points do not constitute a large-scale point cloud, models like DGCNN, PointConv, and PointCNN encounter OOM issues on 4 NVIDIA A10 GPUs. Consequently, we selected PointNet++, PointTransformer, and RandLA-Net as baselines for their efficiency with large-scale point clouds.

\bfsection{Evaluation Metrics}
Due to the significant foreground-background imbalance, the task is defined as binary segmentation, separating the trunk from the spine. Each spine initially receives a unique label during dataset development; however, for experiments, segmentation is binarized to mitigate the imbalance. While these binary results can be further refined into multi-class labels via connected component grouping or clustering (e.g., DBScan), we evaluate model performance using only binary segmentation results to avoid post-processing bias. Specifically, we assess segmentation performance using DSC and IoU for both trunk and spine, with \revised{95\% confidence intervals} for each metric. Spine prediction accuracy is also reported, with an individual spine considered correctly predicted if its Recall exceeds 0.7. \revised{Specifically, for each individual spine $i$, Recall is calculated as $\text{Recall}_i = |\revisedsecond{\hat{y}} \cap y_i| / |y_i|$, where $\revisedsecond{\hat{y}}$ and $y_i$ indicate spine prediction and ground truth labels, respectively.}
All experiments are conducted on 4 NVIDIA A10 GPUs with PyTorch, and detailed settings along with metric tables for each fold are provided in the GitHub repository.

\begin{figure*}[tb]
\centering
\includegraphics[width=\linewidth]{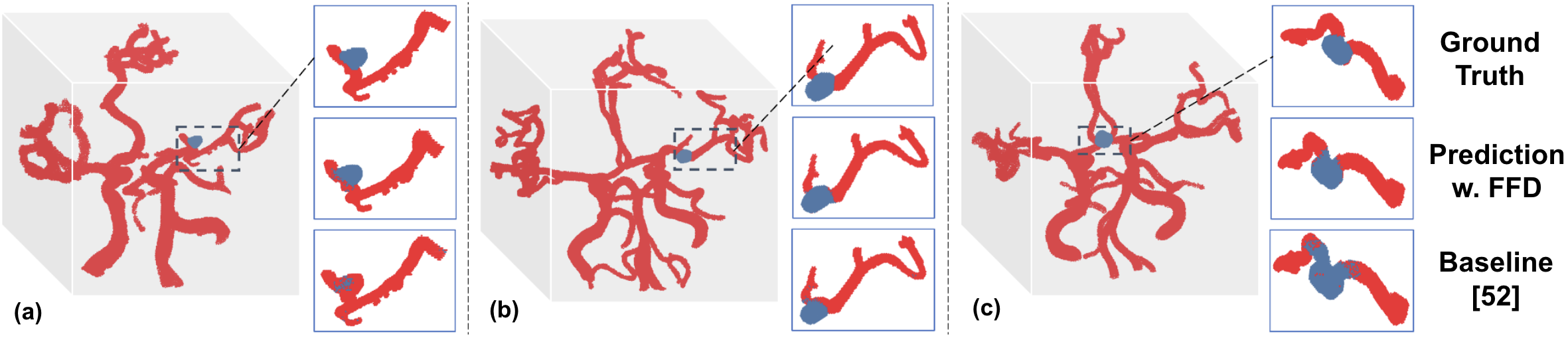}
\caption{\textbf{Additional Evaluation on IntrA.} Visualization results on 3 cases from IntrA dataset are showed. \revised{\textbf{(a):} Aneurysm at vessel endpoint. \textbf{(b):} Saccular aneurysm along vessel segment. \textbf{(c):} Aneurysm at vessel junction. Blue regions indicate aneurysm segmentation. FFD consistently reduces false negatives compared to the baseline, particularly in (a) and (c).}} \label{fig:res_vis_intra}

\end{figure*}

\bfsection{Results and Analysis}
We evaluate the segmentation performance on all three DenSpineEM subsets using models trained on the DenSpineEM-M50 subset.

\itsection{Quantitative Analysis}
We quantitatively evaluate the segmentation performance on the DenSpineEM dataset, as summarized in Table~\ref{tab:model_comparison}. 
\revised{
The experimental results show that PointTransformer with FFD achieves the best performance on binary spine segmentation, with IoU of 90.74\% (95\% CI: 88.55-92.93\%) and DSC of 94.43\% (95\% CI: 92.58-96.28\%). While the baseline PointTransformer achieves slightly higher individual spine accuracy on the M50 dataset (95.67\% vs. 94.67\%), this is likely because high-capacity models can effectively learn complex features directly from raw data when sufficient training examples are available. However, PointTransformer with FFD maintains consistently better performance in zero-shot generalization experiments. On the M10 (cross-region) and H10 (cross-species) subsets, FFD models achieve 95.61\% (95\% CI: 95.50-95.72\%) and 86.63\% (95\% CI: 85.78-87.49\%) DSC scores respectively, outperforming baseline models. PointNet++ with FFD consistently outperforms its baseline counterpart, showing substantial improvements in both IoU (from 69.51\% [95\% CI: 66.76-72.26\%] to 82.60\% [95\% CI: 79.02-86.18\%]) and Dice score (from 80.04\% [95\% CI: 77.53-82.54\%] to 89.28\% [95\% CI: 86.81-91.76\%]) on M50. We note that RandLA-Net models display wider confidence intervals, suggesting potential convergence challenges regardless of whether FFD is applied.}
Overall, adding FFD effectively enhances the models' ability to segment spines accurately, improving both accuracy and generalization.

\itsection{Qualitative Analysis} 
For qualitative analysis, we use predictions from the best-performing model, PointTransformer. \revisedsecond{For visualization clarity, we apply post-processing to assign instance labels: predictions overlapping ground-truth instances inherit their labels, with remaining points labeled based on nearest neighbors.} We visualize two cases from the M50 dataset and one case each from M10 and H10 to evaluate generalization, as shown in Fig.~\ref{fig:res_vis_denspine}. Models with FFD consistently outperform the baseline. On the M50 subset, the baseline predictions contain numerous false negatives, especially on large spines mistaken for trunks ((a), (b)-top), leading to missed spines after clustering. FFD implicitly adds a trunk skeleton prior, alleviating this issue and enhancing model robustness. In generalization tests, the model with FFD maintains robust performance on the M10 subset, while the baseline produces more false positives ((c)-top). For the H10 subset, where dendrites are longer with denser spines, both models' performance degrades. The FFD model includes a few false positives on large spines ((d)-top) and false negatives on small spines ((d)-bottom), whereas the baseline heavily misses many spines with excessive false negatives.


\subsection{Additional Evaluation on Intracranial Aneurysm Segmentation}

\bfsection{Experiment Settings}
We evaluated our method on the IntrA dataset using 5-fold cross-validation on the 103 TOF-MRA samples of the entire artery. The preprocessing pipeline involved voxelizing the surface model using the fast winding number method~\cite{Barill2018FastWN}, \revisedsecond{calibrating the voxel-to-mm scale by matching computed aneurysm sizes to the reported mean of 7.49 mm~\cite{Yang2020ATS},} skeletonizing the artery volume with TEASAR~\cite{TEASAR}, pruning skeleton branches (node degree $<2$ or edge length $<20$ mm\footnote{\revised{The 20 mm threshold was chosen based on the maximum aneurysm size reported in\cite{Yang2020ATS}, ensuring no aneurysm structures are filtered out during skeleton pruning.}}), and cropping the artery into vessel segments \revisedsecond{of 100 mm length to ensure complete aneurysm inclusion with adequate surrounding context}. \revised{We note that brain vessel datasets often contain interruptions in vessel structures due to imaging limitations. Our winding number-based voxelization process naturally helps address this challenge by converting nearby disconnected regions into connected components, enabling subsequent skeletonization to generate continuous curves even when the original data contains minor discontinuities.} We then applied our Frenet-Frame-based transformation and followed the two-step baseline method (detection-segmentation)~\cite{Yang2020ATS}. For fair comparison with the baseline, we first converted voxelized segmentation results back to surface point clouds. \revised{Then we computed the Dice Similarity Coefficient (DSC) on aneurysm prediction to measure the segmentation accuracy.}

\bfsection{Result Analysis}
Our method achieved a DSC \revised{on aneurysm segmentation} of 77.08\% ($\pm$18.75\%), surpassing the previous state-of-the-art performance of 71.79\% ($\pm$29.91\%)~\cite{Yang2020ATS}, which demonstrates both improved accuracy and significantly reduced variability in segmentation results. Fig.~\ref{fig:res_vis_intra} demonstrates the qualitative superiority of applying to FFD over the baseline. In all three cases, our method more accurately delineates aneurysm boundaries (blue regions) within complex arterial structures. 
\revised{
Specifically, for terminal aneurysms located at vessel endpoints (a), the baseline produces significant false negatives, while our FFD-based approach maintains accurate segmentation. For saccular aneurysms appearing as small outpouchings along vessel segments (b), both methods perform well with accurate delineation. In complex cases where aneurysms develop between vessel junctions (c), both methods show some performance degradation, but our FFD approach produces fewer false negatives in the connection regions, demonstrating better robustness to complex vascular geometry.
}

\section{Conclusion}
In this study, we proposed the Frenet–Serret Frame-based Decomposition as an effective solution for accurately segmenting complex 3D curvilinear structures in (bio)medical imaging. By decomposing these structures into globally smooth curves and cylindrical primitives, we achieve reduced representational complexity and enhanced data-efficient learning. Our method demonstrates exceptional cross-region and cross-species generalization on the DenSpineEM dataset, which \revised{is} developed as a comprehensive benchmark for dendritic spine segmentation, achieving high Dice scores in zero-shot segmentation tasks. Additionally, the significant performance improvement on the IntrA dataset underscores its versatility across different medical imaging applications. 

\revisedsecond{While our current evaluation focuses on vessels and neuronal structures, the proposed framework has broader potential for analyzing curvilinear anatomical structures. Our Frenet-Serret frame-based approach could be valuable for tubular organ analysis in various clinical applications, such as gastrointestinal tract assessment for radiotherapy planning, pancreatic duct evaluation, airway tree analysis in pulmonary imaging, and coronary vessel characterization. Recent datasets ~\cite{Lee2024DatasetFG,Ji2022AMOSAL,Yan2024MRISegmentatorAbdomenAF} demonstrate the availability of data for extending our approach to these diverse curvilinear organs, and the geometric decomposition framework could be particularly beneficial for capturing the complex morphological variations inherent in these tubular anatomical structures.}

\bibliographystyle{IEEEtran}
\bibliography{string, reference}

\begin{thebibliography}{10}
\providecommand{\url}[1]{#1}
\csname url@samestyle\endcsname
\providecommand{\newblock}{\relax}
\providecommand{\bibinfo}[2]{#2}
\providecommand{\BIBentrySTDinterwordspacing}{\spaceskip=0pt\relax}
\providecommand{\BIBentryALTinterwordstretchfactor}{4}
\providecommand{\BIBentryALTinterwordspacing}{\spaceskip=\fontdimen2\font plus
\BIBentryALTinterwordstretchfactor\fontdimen3\font minus \fontdimen4\font\relax}
\providecommand{\BIBforeignlanguage}[2]{{%
\expandafter\ifx\csname l@#1\endcsname\relax
\typeout{** WARNING: IEEEtran.bst: No hyphenation pattern has been}%
\typeout{** loaded for the language `#1'. Using the pattern for}%
\typeout{** the default language instead.}%
\else
\language=\csname l@#1\endcsname
\fi
#2}}
\providecommand{\BIBdecl}{\relax}
\BIBdecl

\bibitem{Fornito2015TheCO}
A.~Fornito, A.~Zalesky, and M.~Breakspear, ``The connectomics of brain disorders,'' \emph{Nature Reviews Neuroscience}, vol.~16, pp. 159--172, 2015.

\bibitem{Wu2023MappingON}
J.~Y. Wu, S.-J. Cho, K.~D. Descant, P.~H. Li, A.~Shapson-Coe, M.~Januszewski, D.~R. Berger, C.~Meyer, C.~R. Casingal, A.~Huda, J.~Liu, T.~Ghashghaei, M.~Brenman, M.~Jiang, J.~Scarborough, A.~Pope, V.~Jain, J.~L. Stein, J.~Guo, R.~Yasuda, J.~W. Lichtman, and E.~S. Anton, ``Mapping of neuronal and glial primary cilia contactome and connectome in the human cerebral cortex,'' \emph{Neuron}, vol. 112, pp. 41--55.e3, 2023.

\bibitem{Jumper2021HighlyAP}
J.~M. Jumper, R.~Evans, A.~Pritzel, T.~Green, M.~Figurnov, O.~Ronneberger, K.~Tunyasuvunakool, R.~Bates, A.~Ž{\'i}dek, A.~Potapenko, A.~Bridgland, C.~Meyer, S.~A.~A. Kohl, A.~Ballard, A.~Cowie, B.~Romera-Paredes, S.~Nikolov, R.~Jain, J.~Adler, T.~Back, S.~Petersen, D.~Reiman, E.~Clancy, M.~Zielinski, M.~Steinegger, M.~Pacholska, T.~Berghammer, S.~Bodenstein, D.~Silver, O.~Vinyals, A.~W. Senior, K.~Kavukcuoglu, P.~Kohli, and D.~Hassabis, ``Highly accurate protein structure prediction with alphafold,'' \emph{Nature}, vol. 596, pp. 583 -- 589, 2021.

\bibitem{Abramson2024AccurateSP}
J.~Abramson, J.~Adler, J.~Dunger, R.~Evans, T.~Green, A.~Pritzel, O.~Ronneberger, L.~Willmore, A.~J. Ballard, J.~Bambrick, S.~W. Bodenstein, D.~A. Evans, C.-C. Hung, M.~O’Neill, D.~Reiman, K.~Tunyasuvunakool, Z.~Wu, A.~Žemgulytė, E.~Arvaniti, C.~Beattie, O.~Bertolli, A.~Bridgland, A.~Cherepanov, M.~Congreve, A.~I. Cowen-Rivers, A.~Cowie, M.~Figurnov, F.~B. Fuchs, H.~Gladman, R.~Jain, Y.~A. Khan, C.~M.~R. Low, K.~Perlin, A.~Potapenko, P.~Savy, S.~Singh, A.~Stecula, A.~Thillaisundaram, C.~Tong, S.~Yakneen, E.~D. Zhong, M.~Zielinski, A.~Ž{\'i}dek, V.-. tor Bapst, P.~Kohli, M.~Jaderberg, D.~Hassabis, and J.~M. Jumper, ``Accurate structure prediction of biomolecular interactions with alphafold 3,'' \emph{Nature}, vol. 630, pp. 493 -- 500, 2024.

\bibitem{yang2021ribseg}
J.~Yang, S.~Gu, D.~Wei, H.~Pfister, and B.~Ni, ``Ribseg dataset and strong point cloud baselines for rib segmentation from ct scans,'' in \emph{International Conference on Medical Image Computing and Computer-Assisted Intervention}, 2021.

\bibitem{Jin2022RibSeg}
L.~Jin, S.~Gu, D.~Wei, J.~K. Adhinarta, K.~Kuang, Y.~J. Zhang, H.~Pfister, B.~Ni, J.~Yang, and M.~Li, ``{RibSeg} v2: A large-scale benchmark for rib labeling and anatomical centerline extraction,'' \emph{IEEE Transactions on Medical Imaging}, vol.~43, pp. 570--581, 2024.

\bibitem{Yang2024DeepRF}
J.~Yang, R.~Shi, L.~Jin, X.~Huang, K.~Kuang, D.~Wei, S.~Gu, J.~Liu, P.~Liu, Z.~Chai, Y.~Xiao, H.~Chen, L.~Xu, B.~Du, X.~Yan, H.~Tang, A.~Alessio, G.~Holste, J.~Zhang, X.~Wang, J.~He, L.~Che, H.~Pfister, M.~Li, and B.~Ni, ``Deep rib fracture instance segmentation and classification from ct on the ribfrac challenge,'' \emph{ArXiv}, vol. abs/2402.09372, 2024.

\bibitem{Yang2020IntrA3I}
X.~Yang, D.~Xia, T.~Kin, and T.~Igarashi, ``Intra: 3d intracranial aneurysm dataset for deep learning,'' \emph{2020 IEEE/CVF Conference on Computer Vision and Pattern Recognition (CVPR)}, pp. 2653--2663, 2020.

\bibitem{Isensee2020nnUNetAS}
F.~Isensee, P.~F. Jaeger, S.~A.~A. Kohl, J.~Petersen, and K.~H. Maier-Hein, ``nnu-net: a self-configuring method for deep learning-based biomedical image segmentation,'' \emph{Nature Methods}, vol.~18, pp. 203 -- 211, 2020.

\bibitem{Kashiwagi2019ComputationalGA}
Y.~Kashiwagi, T.~Higashi, K.~Obashi, Y.~Sato, N.~H. Komiyama, S.~G.~N. Grant, and S.~Okabe, ``Computational geometry analysis of dendritic spines by structured illumination microscopy,'' \emph{Nature Communications}, vol.~10, 2019.

\bibitem{Boros2017DendriticSP}
B.~D. Boros, K.~M. Greathouse, E.~G. Gentry, K.~Curtis, E.~L. Birchall, M.~Gearing, and J.~H. Herskowitz, ``Dendritic spines provide cognitive resilience against alzheimer's disease,'' \emph{Annals of Neurology}, vol.~82, 2017.

\bibitem{Qian2023BiomimeticIN}
K.~Qian, A.~S. Liao, S.~Gu, V.~A. Webster-Wood, and Y.~J. Zhang, ``Biomimetic {IGA} neuron growth modeling with neurite morphometric features and {CNN}-based prediction,'' \emph{Computer Methods in Applied Mechanics and Engineering}, vol. 417, p. 116213, 2023.

\bibitem{VidaurreGallart2022ADL}
I.~Vidaurre-Gallart, I.~Fernaud‐Espinosa, N.~Cosmin-Toader, L.~Talavera-Mart{\'i}nez, M.~Martin-Abadal, R.~Benavides-Piccione, Y.~Gonz{\'a}lez-Cid, L.~Pastor, J.~DeFelipe, and M.~Garcia-Lorenzo, ``A deep learning-based workflow for dendritic spine segmentation,'' \emph{Frontiers in Neuroanatomy}, vol.~16, 2022.

\bibitem{Erdil2015AJC}
E.~Erdil, A.~O. Argunsah, T.~Tasdizen, D.~{\"U}nay, and M.~Çetin, ``A joint classification and segmentation approach for dendritic spine segmentation in 2-photon microscopy images,'' \emph{2015 IEEE 12th International Symposium on Biomedical Imaging (ISBI)}, pp. 797--800, 2015.

\bibitem{Basu20193dSpAnAI}
S.~Basu, N.~Das, E.~Baczynska, M.~Bijata, A.~Zeug, D.~M. Plewczynski, P.~K. Saha, E.~G. Ponimaskin, and J.~Włodarczyk, ``3dspan: an interactive software for 3d segmentation and analysis of dendritic spines,'' \emph{bioRxiv}, 2019.

\bibitem{Lesage2009ARO}
D.~Lesage, E.~D. Angelini, I.~Bloch, and G.~Funka-Lea, ``A review of 3d vessel lumen segmentation techniques: Models, features and extraction schemes,'' \emph{Medical image analysis}, vol. 13 6, pp. 819--45, 2009.

\bibitem{Acciai2016AutomatedNT}
L.~Acciai, P.~Soda, and G.~Iannello, ``Automated neuron tracing methods: An updated account,'' \emph{Neuroinformatics}, vol.~14, pp. 353 -- 367, 2016.

\bibitem{Lo2012ExtractionOA}
P.~Lo, B.~van Ginneken, J.~M. Reinhardt, T.~Yavarna, P.~A. de~Jong, B.~Irving, C.~I. Fetita, M.~Ortner, R.~Pinho, J.~Sijbers, M.~Feuerstein, A.~Fabijańska, C.~Bauer, R.~R. Beichel, C.~S. Mendoza, R.~Wiemker, J.~Lee, A.~P. Reeves, S.~Born, O.~Weinheimer, E.~M. van Rikxoort, J.~Tschirren, K.~Mori, B.~Odry, D.~P. Naidich, I.~Hartmann, E.~A. Hoffman, M.~Prokop, J.~J.~H. Pedersen, and M.~de~Bruijne, ``Extraction of airways from ct (exact'09),'' \emph{IEEE Transactions on Medical Imaging}, vol.~31, pp. 2093--2107, 2012.

\bibitem{Frangi1998MuliscaleVE}
A.~F. Frangi, W.~J. Niessen, K.~L. Vincken, and M.~A. Viergever, ``Muliscale vessel enhancement filtering,'' in \emph{International Conference on Medical Image Computing and Computer-Assisted Intervention}, 1998.

\bibitem{Sato1998ThreedimensionalML}
Y.~Sato, S.~Nakajima, N.~Shiraga, H.~Atsumi, S.~Yoshida, T.~Koller, G.~Gerig, and R.~Kikinis, ``Three-dimensional multi-scale line filter for segmentation and visualization of curvilinear structures in medical images,'' \emph{Medical image analysis}, vol. 2 2, pp. 143--68, 1998.

\bibitem{Sironi2016MultiscaleCD}
A.~Sironi, E.~T{\"u}retken, V.~Lepetit, and P.~V. Fua, ``Multiscale centerline detection,'' \emph{IEEE Transactions on Pattern Analysis and Machine Intelligence}, vol.~38, pp. 1327--1341, 2016.

\bibitem{Tetteh2018DeepVesselNetVS}
G.~Tetteh, V.~Efremov, N.~D. Forkert, M.~Schneider, J.~S. Kirschke, B.~Weber, C.~Zimmer, M.~Piraud, and B.~H. Menze, ``Deepvesselnet: Vessel segmentation, centerline prediction, and bifurcation detection in 3-d angiographic volumes,'' \emph{Frontiers in Neuroscience}, vol.~14, 2018.

\bibitem{Litjens2017ASO}
G.~J.~S. Litjens, T.~Kooi, B.~E. Bejnordi, A.~A.~A. Setio, F.~Ciompi, M.~Ghafoorian, J.~van~der Laak, B.~van Ginneken, and C.~I. S{\'a}nchez, ``A survey on deep learning in medical image analysis,'' \emph{Medical image analysis}, vol.~42, pp. 60--88, 2017.

\bibitem{Shit2020clDiceA}
S.~Shit, J.~C. Paetzold, A.~K. Sekuboyina, I.~Ezhov, A.~Unger, A.~Zhylka, J.~P.~W. Pluim, U.~Bauer, and B.~H. Menze, ``cldice - a novel topology-preserving loss function for tubular structure segmentation,'' \emph{2021 IEEE/CVF Conference on Computer Vision and Pattern Recognition (CVPR)}, pp. 16\,555--16\,564, 2020.

\bibitem{Kervadec2021BeyondPS}
H.~Kervadec, H.~Bahig, L.~L{\'e}tourneau-Guillon, J.~Dolz, and I.~B. Ayed, ``Beyond pixel-wise supervision for segmentation: A few global shape descriptors might be surprisingly good!'' in \emph{International Conference on Medical Imaging with Deep Learning}, 2021.

\bibitem{Yao2024AASegAG}
L.~Yao, D.~Chen, X.~Zhao, M.~Fei, Z.~Song, Z.~Xue, Y.~Zhan, B.~Song, F.~Shi, Q.~Wang, and D.~Shen, ``Aaseg: Artery-aware global-to-local framework for aneurysm segmentation in head and neck cta images,'' \emph{IEEE Transactions on Medical Imaging}, vol.~44, pp. 1273--1283, 2024.

\bibitem{Wang2022BowelNetJS}
C.~Wang, Z.~Cui, J.~Yang, M.~Han, G.~Carneiro, and D.~Shen, ``Bowelnet: Joint semantic-geometric ensemble learning for bowel segmentation from both partially and fully labeled ct images,'' \emph{IEEE Transactions on Medical Imaging}, vol.~42, pp. 1225--1236, 2022.

\bibitem{Hu2021TopologyAwareSU}
X.~Hu, Y.~Wang, F.~Li, D.~Samaras, and C.~Chen, ``Topology-aware segmentation using discrete morse theory,'' \emph{ArXiv}, vol. abs/2103.09992, 2021.

\bibitem{Hu2019TopologyPreservingDI}
X.~Hu, F.~Li, D.~Samaras, and C.~Chen, ``Topology-preserving deep image segmentation,'' \emph{ArXiv}, vol. abs/1906.05404, 2019.

\bibitem{ravishankar2017learning}
H.~Ravishankar, R.~Venkataramani, S.~R. Thiruvenkadam, P.~Sudhakar, and V.~Vaidya, ``Learning and incorporating shape models for semantic segmentation,'' in \emph{International Conference on Medical Image Computing and Computer-Assisted Intervention}, 2017.

\bibitem{wang2020deep}
Y.~Wang, X.~Wei, F.~Liu, J.~Chen, Y.~Zhou, W.~Shen, E.~K. Fishman, and A.~L. Yuille, ``Deep distance transform for tubular structure segmentation in ct scans,'' 2019, pp. 3832--3841.

\bibitem{cciccek20163d}
{\"O}.~Çiçek, A.~Abdulkadir, S.~S. Lienkamp, T.~Brox, and O.~Ronneberger, ``3d u-net: Learning dense volumetric segmentation from sparse annotation,'' in \emph{International Conference on Medical Image Computing and Computer-Assisted Intervention}, 2016.

\bibitem{yang2021reinventing}
J.~Yang, X.~Huang, Y.~He, J.~Xu, C.~Yang, G.~Xu, and B.~Ni, ``Reinventing 2d convolutions for 3d images,'' \emph{IEEE Journal of Biomedical and Health Informatics}, vol.~25, pp. 3009--3018, 2019.

\bibitem{qi2017pointnet}
C.~Qi, H.~Su, K.~Mo, and L.~J. Guibas, ``Pointnet: Deep learning on point sets for 3d classification and segmentation,'' 2016, pp. 77--85.

\bibitem{ho2021point}
N.-V. Ho, T.~H. Nguyen, G.-H. Diep, N.~T.~H. Le, and B.-S. Hua, ``Point-unet: A context-aware point-based neural network for volumetric segmentation,'' in \emph{International Conference on Medical Image Computing and Computer-Assisted Intervention}, 2022.

\bibitem{Liu2022EdgeOrientedPT}
Y.~Liu, J.~Liu, and Y.~Yuan, ``Edge-oriented point-cloud transformer for 3d intracranial aneurysm segmentation,'' in \emph{International Conference on Medical Image Computing and Computer-Assisted Intervention}, 2022.

\bibitem{Xie2023EfficientAL}
K.~Xie, J.~Yang, D.~Wei, Z.~Weng, and P.~Fua, ``Efficient anatomical labeling of pulmonary tree structures via deep point-graph representation-based implicit fields,'' \emph{Medical image analysis}, vol.~99, p. 103367, 2023.

\bibitem{Nimchinsky2002StructureAF}
E.~A. Nimchinsky, B.~L. Sabatini, and K.~Svoboda, ``Structure and function of dendritic spines.'' \emph{Annual review of physiology}, vol.~64, pp. 313--53, 2002.

\bibitem{xiao2018automated}
X.~Xiao, M.~Djurisic, A.~Hoogi, R.~W. Sapp, and D.~Rubin, ``Automated dendritic spine detection using convolutional neural networks on maximum intensity projected microscopic volumes,'' \emph{Journal of Neuroscience Methods}, vol. 309, pp. 25--34, 2018.

\bibitem{choi2021dxplorer}
J.~Choi, S.-E. Lee, Y.~Lee, E.~Cho, S.~Chang, and W.-K. Jeong, ``Dxplorer: A unified visualization framework for interactive dendritic spine analysis using 3d morphological features,'' \emph{IEEE Transactions on Visualization and Computer Graphics}, vol.~PP, pp. 1--1, 2021.

\bibitem{wildenberg2021large}
G.~Wildenberg, H.~Li, G.~Badalamente, T.~D. Uram, N.~J. Ferrier, and N.~Kasthuri, ``Large-scale dendritic spine extraction and analysis through petascale computing,'' \emph{bioRxiv}, 2021.

\bibitem{dorkenwald2019binary}
S.~Dorkenwald, N.~L. Turner, T.~Macrina, K.~Lee, R.~Lu, J.~Wu, A.~L. Bodor, A.~A. Bleckert, D.~Brittain, N.~Kemnitz, W.~M. Silversmith, D.~Ih, J.~Zung, A.~Zlateski, I.~Tartavull, S.~chieh Yu, S.~Popovych, W.~Wong, M.~A. Castro, C.~S. Jordan, A.~M. Wilson, E.~Froudarakis, J.~Buchanan, M.~M. Takeno, R.~Torres, G.~Mahalingam, F.~Collman, C.~M. Schneider-Mizell, D.~Bumbarger, Y.~Li, L.~Becker, S.~K. Suckow, J.~Reimer, A.~S. Tolias, N.~M. da~Costa, R.~C. Reid, and H.~S. Seung, ``Binary and analog variation of synapses between cortical pyramidal neurons,'' \emph{eLife}, vol.~11, 2019.

\bibitem{Piaggio1952DifferentialGO}
K.~Tapp, ``Differential geometry of curves and surfaces,'' \emph{Nature}, vol. 169, pp. 560--560, 1952.

\bibitem{Bishop1975ThereIM}
R.~Bishop, ``There is more than one way to frame a curve,'' \emph{American Mathematical Monthly}, vol.~82, pp. 246--251, 1975.

\bibitem{Huang2023TrajectoryPI}
J.~Huang, Z.~He, Y.~Arakawa, and B.~Dawton, ``Trajectory planning in frenet frame via multi-objective optimization,'' \emph{IEEE Access}, vol.~11, pp. 70\,764--70\,777, 2023.

\bibitem{Kseolu2023InvolutiveSS}
G.~K{\"o}seoğlu and M.~K. Bilici, ``Involutive sweeping surfaces with frenet frame in euclidean 3-space,'' \emph{Heliyon}, vol.~9, 2023.

\bibitem{Hanson1995QuaternionFA}
A.~J. Hanson and H.~Ma, ``Quaternion frame approach to streamline visualization,'' \emph{IEEE Trans. Vis. Comput. Graph.}, vol.~1, pp. 164--174, 1995.

\bibitem{Pottmann1998ContributionsTM}
H.~Pottmann and M.~G. Wagner, ``Contributions to motion based surface design,'' \emph{Int. J. Shape Model.}, vol.~4, pp. 183--196, 1998.

\bibitem{Hu2011DiscreteFF}
S.~Hu, M.~Lundgren, and A.~J. Niemi, ``Discrete frenet frame, inflection point solitons, and curve visualization with applications to folded proteins.'' \emph{Physical review. E, Statistical, nonlinear, and soft matter physics}, vol. 83 6 Pt 1, p. 061908, 2011.

\bibitem{TEASAR}
M.~Sato, I.~Bitter, M.~A. Bender, A.~E. Kaufman, and M.~Nakajima, ``{TEASAR}: tree-structure extraction algorithm for accurate and robust skeletons,'' 2000, pp. 281--449.

\bibitem{Bitter2001PenalizedDistanceVS}
I.~Bitter, A.~E. Kaufman, and M.~Sato, ``Penalized-distance volumetric skeleton algorithm,'' \emph{IEEE Trans. Vis. Comput. Graph.}, vol.~7, pp. 195--206, 2001.

\bibitem{Yang2020ATS}
X.~Yang, D.~Xia, T.~Kin, and T.~Igarashi, ``A two-step surface-based 3d deep learning pipeline for segmentation of intracranial aneurysms,'' \emph{Computational Visual Media}, vol.~9, pp. 57--69, 2020.

\bibitem{Huang2013L1medialSO}
H.~Huang, S.~Wu, D.~Cohen-Or, M.~Gong, H.~Zhang, G.~Li, and B.~Chen, ``L1-medial skeleton of point cloud,'' \emph{ACM Transactions on Graphics (TOG)}, vol.~32, pp. 1 -- 8, 2013.

\bibitem{Chrzeszczyk2011MatrixC}
A.~Chrzeszczyk and J.~Kochanowski, ``Matrix computations,'' in \emph{Encyclopedia of Parallel Computing}, 2011.

\bibitem{ofer2021ultrastructural}
N.~Ofer, D.~R. Berger, N.~Kasthuri, J.~W. Lichtman, and R.~Yuste, ``Ultrastructural analysis of dendritic spine necks reveals a continuum of spine morphologies,'' \emph{bioRxiv}, 2021.

\bibitem{Kasthuri2015}
N.~Kasthuri, K.~J. Hayworth, D.~R. Berger, R.~L. Schalek, J.~A. Conchello, S.~Knowles-Barley, D.~Lee, A.~V{\'a}zquez-Reina, V.~Kaynig, T.~R. Jones, M.~Roberts, J.~L. Morgan, J.~C. Tapia, H.~S. Seung, W.~R.~G. Roncal, J.~T. Vogelstein, R.~C. Burns, D.~L. Sussman, C.~E. Priebe, H.~Pfister, and J.~W. Lichtman, ``Saturated reconstruction of a volume of neocortex,'' \emph{Cell}, vol. 162, pp. 648--661, 2015.

\bibitem{wei2021axonem}
D.~Wei, K.~Lee, H.~Li, R.~Lu, J.~A. Bae, Z.~Liu, L.~Zhang, M.~dos Santos, Z.~Lin, T.~D. Uram, X.~Wang, I.~Arganda-Carreras, B.~Matejek, N.~Kasthuri, J.~W. Lichtman, and H.~Pfister, ``Axonem dataset: 3d axon instance segmentation of brain cortical regions,'' vol. abs/2107.05451, 2021.

\bibitem{berger2018vast}
D.~R. Berger, H.~S. Seung, and J.~W. Lichtman, ``Vast (volume annotation and segmentation tool): Efficient manual and semi-automatic labeling of large 3d image stacks,'' \emph{Frontiers in Neural Circuits}, vol.~12, 2018.

\bibitem{ng}
\BIBentryALTinterwordspacing
J.~Maitin-Shepard, ``Neuroglancer: Webgl-based viewer for volumetric data,'' 2017. [Online]. Available: \url{https://github.com/google/neuroglancer}
\BIBentrySTDinterwordspacing

\bibitem{Li2018PointCNNCO}
Y.~Li, R.~Bu, M.~Sun, W.~Wu, X.~Di, and B.~Chen, ``Pointcnn: Convolution on x-transformed points,'' in \emph{Neural Information Processing Systems}, 2018.

\bibitem{qi2017pointnet++}
C.~Qi, L.~Yi, H.~Su, and L.~J. Guibas, ``Pointnet++: Deep hierarchical feature learning on point sets in a metric space,'' \emph{ArXiv}, vol. abs/1706.02413, 2017.

\bibitem{Wang2018DynamicGC}
Y.~Wang, Y.~Sun, Z.~Liu, S.~E. Sarma, M.~M. Bronstein, and J.~M. Solomon, ``Dynamic graph cnn for learning on point clouds,'' \emph{ACM Transactions on Graphics (TOG)}, vol.~38, pp. 1 -- 12, 2018.

\bibitem{Yu20213DMP}
J.~Yu, C.~Zhang, H.~Wang, D.~Zhang, Y.~Song, T.~Xiang, D.~Liu, and W.~T. Cai, ``3d medical point transformer: Introducing convolution to attention networks for medical point cloud analysis,'' \emph{ArXiv}, vol. abs/2112.04863, 2021.

\bibitem{Hu2019RandLANetES}
Q.~Hu, B.~Yang, L.~Xie, S.~Rosa, Y.~Guo, Z.~Wang, A.~Trigoni, and A.~Markham, ``Randla-net: Efficient semantic segmentation of large-scale point clouds,'' \emph{2020 IEEE/CVF Conference on Computer Vision and Pattern Recognition (CVPR)}, pp. 11\,105--11\,114, 2019.

\bibitem{Wu2023PointTV}
X.~Wu, L.~Jiang, P.-S. Wang, Z.~Liu, X.~Liu, Y.~Qiao, W.~Ouyang, T.~He, and H.~Zhao, ``Point transformer v3: Simpler, faster, stronger,'' \emph{2024 IEEE/CVF Conference on Computer Vision and Pattern Recognition (CVPR)}, pp. 4840--4851, 2023.

\bibitem{Barill2018FastWN}
G.~Barill, N.~G. Dickson, R.~M. Schmidt, D.~I.~W. Levin, and A.~Jacobson, ``Fast winding numbers for soups and clouds,'' \emph{ACM Transactions on Graphics (TOG)}, vol.~37, pp. 1 -- 12, 2018.

\bibitem{Lee2024DatasetFG}
S.~L. Lee, P.~Yadav, Y.~Li, J.~J. Meudt, J.~Strang, D.~J. Hebel, A.~Alfson, S.~J. Olson, T.~R. Kruser, J.~B. Smilowitz, K.~N. Borchert, B.~Loritz, L.~A. Gharzai, S.~Karimpour, J.~Bayouth, and M.~F. Bassetti, ``Dataset for gastrointestinal tract segmentation on serial mris for abdominal tumor radiotherapy,'' \emph{Data in Brief}, vol.~57, 2024.

\bibitem{Ji2022AMOSAL}
Y.~Ji, H.~Bai, J.~Yang, C.~Ge, Y.~Zhu, R.~Zhang, Z.~Li, L.~Zhang, W.~Ma, X.~Wan, and P.~Luo, ``Amos: A large-scale abdominal multi-organ benchmark for versatile medical image segmentation,'' \emph{ArXiv}, vol. abs/2206.08023, 2022.

\bibitem{Yan2024MRISegmentatorAbdomenAF}
Z.~Yan, T.~Mathai, P.~Mukherjee, B.~Khoury, B.~Kim, B.~Hou, N.~Rabbee, and R.~M. Summers, ``Mrisegmentator-abdomen: A fully automated multi-organ and structure segmentation tool for t1-weighted abdominal mri,'' \emph{ArXiv}, 2024.

\end{thebibliography}

\end{document}